\documentclass{article}

\usepackage[arXiv]{icml2020}

\icmltitlerunning{Probabilistic Decoupling of Labels in Classification}

\usepackage[latin1]{inputenc} 
\usepackage[T1]{fontenc}    
\usepackage{hyperref}       
\usepackage{url}            
\usepackage{booktabs}       
\usepackage{amsfonts}       
\usepackage{nicefrac}       
\usepackage{microtype}      

\usepackage[dvipsnames]{xcolor}

\usepackage{hyperref}
\usepackage{url}

\hypersetup{hidelinks}
\usepackage{calc}
\usepackage{lipsum}

\usepackage{mathtools}
\usepackage{booktabs}       
\usepackage{amsfonts}       
\usepackage{nicefrac}       
\usepackage{microtype}      

\usepackage{amsmath}
\usepackage{amssymb}
\usepackage{mathrsfs}
\usepackage{microtype}     
\usepackage{accents}
\usepackage{setspace}		
\usepackage{bm}			    
\usepackage{dashrule}		
\usepackage{varioref}		
\usepackage{subcaption}		
\usepackage{caption}		
\usepackage{wrapfig}		
\usepackage{color}			
\usepackage{multicol}		
\usepackage[super]{nth}		
\usepackage{bold-extra} 	
\usepackage{graphbox}		

\usepackage{longtable}		
\usepackage{stmaryrd}		
\usepackage{listings}		
\usepackage{pdfpages}		
\usepackage{sidecap}		
\sidecaptionvpos{figure}{c}
\usepackage{multirow}		
\usepackage{framed}			
\usepackage{units}			
\usepackage{bold-extra}		
\usepackage[export]{adjustbox} 
\usepackage{booktabs}		
\usepackage{needspace}		
\usepackage{relsize}		
\usepackage{pifont} 		
\usepackage{threeparttable}
\usepackage[counterclockwise, figuresleft]{rotating}		
\usepackage{upgreek}  		
\usepackage{tcolorbox}
\usepackage{setspace}
\usepackage{graphicx}
\usepackage{subfiles}
\usepackage{varwidth}
\usepackage{lscape}  
\usepackage[bottom]{footmisc}  
\usepackage{mdframed}  
\usepackage{enumitem}

\definecolor{shadecolor}{RGB}{144, 238, 144}

\makeatletter
\newsavebox\myboxA
\newsavebox\myboxB
\newlength\mylenA
\newcommand*\xbar[2][0.75]{%
	\sbox{\myboxA}{$\m@th#2$}%
	\setbox\myboxB\null
	\ht\myboxB=\ht\myboxA%
	\dp\myboxB=\dp\myboxA%
	\wd\myboxB=#1\wd\myboxA
	\sbox\myboxB{$\m@th\overline{\copy\myboxB}$}
	\setlength\mylenA{\the\wd\myboxA}
	\addtolength\mylenA{-\the\wd\myboxB}%
	\ifdim\wd\myboxB<\wd\myboxA%
	\rlap{\hskip 0.5\mylenA\usebox\myboxB}{\usebox\myboxA}%
	\else
	\hskip -0.5\mylenA\rlap{\usebox\myboxA}{\hskip 0.5\mylenA\usebox\myboxB}%
	\fi}
\makeatother

\renewcommand{\bar}[1]{\overline{#1}}

\newcommand{\temp}{def}

\DeclareMathOperator{\dirichlet}{Dir}

\DeclareMathOperator{\categorical}{Categorical}
\DeclareMathOperator{\Dirp}{DirP}

\renewcommand{\digamma}{\psi}
\newcommand{\trigamma}{\psi_1}

\setlist{itemsep=0ex, topsep=0.7ex}

\definecolor{sRed}{cmyk}{0, 0.91, 0.72, 0.23}
\definecolor{sGray}{cmyk}{0, 0, 0, 0.56}

\definecolor{s_sek_1}{cmyk}{0, 0.5, 1.0, 0} 
\definecolor{s_sek_2}{cmyk}{0, 0.75, 1.0, 0} 
\definecolor{s_sek_3}{cmyk}{0, 1, 1, 0} 
\definecolor{s_sek_4}{cmyk}{0, 1, 1, 0.5} 
\definecolor{s_sek_5}{cmyk}{0, 1, 0, 0} 
\definecolor{s_sek_6}{cmyk}{0.25, 1, 0, 0} 
\definecolor{s_sek_7}{cmyk}{0.5, 1, 0, 0} 
\definecolor{s_sek_8}{cmyk}{0.75, 1, 0, 0} 
\definecolor{s_sek_9}{cmyk}{0.75, 0.75, 0, 0} 
\definecolor{s_sek_10}{cmyk}{0.5, 0, 0, 0} 
\definecolor{s_sek_11}{cmyk}{0.25, 0, 1, 0} 
\definecolor{s_sek_12}{cmyk}{0, 0.25, 1.0, 0} 
\definecolor{s_sek_13}{cmyk}{0.75, 0.5, 0, 0} 
\definecolor{s_sek_14}{cmyk}{0.5, 0, 1, 0} 

\colorlet{s_sek_orange}{s_sek_1}
\colorlet{s_sek_orange_red}{s_sek_2}
\colorlet{s_sek_red}{s_sek_3}
\colorlet{s_sek_dark_red}{s_sek_4}
\colorlet{s_sek_light_pink}{s_sek_5}
\colorlet{s_sek_pink}{s_sek_6}
\colorlet{s_sek_purple}{s_sek_7}
\colorlet{s_sek_purple_blue}{s_sek_8}
\colorlet{s_sek_blue}{s_sek_9}
\colorlet{s_sek_turqoise}{s_sek_10}
\colorlet{s_sek_green_yellow}{s_sek_11}
\colorlet{s_sek_yellow}{s_sek_12}
\colorlet{s_sek_light_blue}{s_sek_13}
\colorlet{s_sek_green}{s_sek_14}

\usepackage{titlesec}		

\usepackage[textsize=scriptsize, textwidth=3.4cm]{todonotes}		
\presetkeys{todonotes}{backgroundcolor=green, linecolor=black}{}

\usepackage{enumitem}

\usepackage{array}			
\newcounter{rowcount}

\usepackage{array} 		
\newcolumntype{L}[2] 	
{>{\raggedright\let\newline\\\arraybackslash\hspace{0pt}}#1{#2}}
\newcolumntype{C}[2]	
{>{\centering\let\newline\\\arraybackslash\hspace{0pt}}#1{#2}}
\newcolumntype{R}[2]	
{>{\raggedleft\let\newline\\\arraybackslash\hspace{0pt}}#1{#2}}

\usepackage{amsmath,amsfonts,bm,upgreek,cprotect}

\SetSymbolFont{largesymbols}{bold}{OMX}{txex}{b}{n}  

\DeclareMathOperator{\lagrange}{\mathscr{L}}

\newcommand{\appropto}{\mathrel{\vcenter{
	\offinterlineskip\halign{\hfil$##$\cr
		\propto\cr\noalign{\kern2pt}\sim\cr\noalign{\kern-2pt}}}}}

\DeclareMathOperator{\e}{e}                                     

\newcommand{\diff}[2][]{\frac{\text{d}#1}{\text{d}#2}}          
\renewcommand{\d}{\:\text{d}}                                   

\newcommand{\N}{\mathbb{N}}                                     
\newcommand{\R}{\mathbb{R}}                                     

\DeclareMathOperator{\E}{\mathbb{E}}                            
\DeclareMathOperator{\bln}{\textbf{ln}}                         
\DeclareMathOperator{\var}{var}                                 
\DeclareMathOperator*{\amax}{arg\,max}                          

\newcommand{\given}{\: | \:}

\DeclareMathAlphabet{\mathsfit}{\encodingdefault}{\sfdefault}{m}{sl}
\SetMathAlphabet{\mathsfit}{bold}{\encodingdefault}{\sfdefault}{bx}{n}
\newcommand{\tens}[1]{\bm{\mathsfit{#1}}}

\def\vp{{\bf{p}}}

\def\vr{{\bf{r}}}
\def\vs{{\bf{s}}}

\def\vv{{\bf{v}}}

\def\vx{{\bf{x}}}
\def\vy{{\bf{y}}}
\def\vz{{\bf{z}}}

\def\vone{{\bm{1}}}

\newcommand{\valpha}{{\bm{\upalpha}}}
\newcommand{\vbeta}{{\bm{\upbeta}}}
\newcommand{\vgamma}{{\bm{\upgamma}}}

\newcommand{\vlambda}{{\bm{\uplambda}}}

\newcommand{\vpi}{{\bm{\uppi}}}

\def\eva{\text{a}}

\def\evc{\text{c}}

\def\evp{\text{p}}

\def\evr{\text{r}}

\def\evw{\text{w}}
\def\evx{\text{x}}

\newcommand{\evalpha}{{\upalpha}}
\newcommand{\evbeta}{{\upbeta}}
\newcommand{\evgamma}{{\upgamma}}

\newcommand{\evlambda}{{\uplambda}}

\newcommand{\evpi}{{\uppi}}

\def\mA{{\bf{A}}}

\def\mS{{\bf{S}}}
\def\mT{{\bf{T}}}

\def\mW{{\bf{W}}}
\def\mX{{\bf{X}}}
\def\mY{{\bf{Y}}}

\newcommand{\mTheta}{{\bf{\Theta}}}

\newcommand{\mPi}{{\bf{\Pi}}}

\def\emA{\text{A}}

\def\emS{\text{S}}
\def\emT{\text{T}}

\def\emX{\text{X}}
\def\emY{\text{Y}}

\newcommand{\emTheta}{{\Theta}}

\newcommand{\emPi}{{\Pi}}

\def\tS{{\tens{S}}}

\newcommand{\etens}[1]{\mathsfit{#1}}

\def\etS{{\etens{S}}}

\renewcommand{\d}{\:\text{d}}

\begin{document}

\twocolumn[
\icmltitle{Probabilistic Decoupling of Labels in Classification}




\begin{icmlauthorlist}
\icmlauthor{Jeppe N{\o}rregaard}{dtu,git}
\icmlauthor{Lars Kai Hansen}{dtu}
\end{icmlauthorlist}

\icmlaffiliation{dtu}{DTU Compute, Technical University of Denmark, Denmark}
\icmlaffiliation{git}{\url{github.com/NorthGuard/}}

\icmlcorrespondingauthor{Jeppe N{\o}rregaard}{jepno@dtu.dk}

\icmlkeywords{Machine Learning, ICML}

\vskip 0.3in
]



\printAffiliationsAndNotice{}  
\raggedbottom


\begin{abstract}
	In this paper we develop a principled, probabilistic, unified approach to non-standard classification tasks, such as semi-supervised, positive-unlabelled, multi-positive-unlabelled and noisy-label learning. We train a classifier on the given labels to predict the \textit{label}-distribution. We then infer the underlying \textit{class}-distributions by variationally optimizing a model of label-class transitions. 
\end{abstract}

\section{Introduction}

Label uncertainty and availability are crucial aspects of supervised classification, and has led to a range of non-standard classification problems such as semi-supervised, noisy-label, positive-unlabelled and multi-positive-unlabelled learning. Semi-supervised learning has acquired enormous attention, but noisy-label learning, and especially positive-unlabelled and multi-positive-unlabelled learning, have made less progress. While some methodologies have been applied to both semi-supervised and noisy-label learning, there have been limited attempts at universal methods for these non-standard classification settings. Furthermore many methodologies are architectural and heuristic, and lack theoretical foundation. \\

In this paper we propose a principled, unified approach to non-standard classification in general. It allows for encoding of prior knowledge, possible inference and updates for these priors, and generalizes the theoretical foundation of positive-unlabelled learning laid out by Elkan and Noto \cite{elkan_learning_2008}.

\section{Related Works}
Semi-supervised learning refers to classification problems where some of the training samples are labelled and others are not. In this setting we hope to utilize the unlabelled data to improve performance compared to only using the labelled part. A common approach, called transductive semi-supervised learning \cite{zhu_introduction_2009, triguero_self-labeled_2015}, is to attempt to predict labels on the unlabelled dataset and then use the combined dataset to train final models. One transductive method is self-training in which a model switches between training and relabelling its own training data. In Yarowsky's algorithm \cite{yarowsky_unsupervised_1995, tanha_semi-supervised_2017} we include all samples with the probability of a class over some threshold. Alternative approaches rank samples according to probability of their most probable classes, and select top-$k$ samples for the classes \cite{zhang_enhanced_2016}, while possibly maintaining class-balance. SETRED \cite{li_setred_2005} is a model that combines any classifier with graph-based method and use the combined prediction to ensure good labels. SETRED has since publication been shown to work really well \cite{triguero_self-labeled_2015}. The method is similar to co-training methods, in which two (or more) views of the same data (same samples but different features) are used to train two (or more) models, and the combined performance is then used \cite{blum_combining_1998, goldman_enhancing_2009}. MixMatch \cite{berthelot_mixmatch_2019} is a method that uses a model to predict labels on unlabelled data (which is improved using data augmentation and sharpening of labels), and uses MixUp \cite{zhang_mixup_2018} to encourage convex behaviour between samples (MixUp simulates examples as linear combinations of samples). Alternative approaches to semi-supervised learning include EM-approaches \cite{miller_mixture_1997, nigam_semi_supervised_2006}, where the true labels of unlabelled data are considered latent variables to be optimized, graph-based methods such as label-propagation \cite{zhu_learning_2009} and nearest neighbour methods \cite{wang_semi-supervised_2010}, and some dedicated semi-supervised models such as the $\text{S}^3\text{VM}$ \cite{bennett_semi_supervised_1999}, Gaussian processes  \cite{lawrence_semi-supervised_2005} and deep learning models trained with entropy regularization  \cite{grandvalet_semi-supervised_2013}. \\

Positive-unlabelled learning is a variant of semi-supervised learning in which we attempt to predict two classes, but only have labels for one of them (the positive class), as well as unlabelled data. Elkan and Noto \cite{elkan_learning_2008} created the theoretical foundation for positive-unlabelled learning (which we generalize). Their method requires estimating the class prior which has since then been improved by Du Plessis \cite{du_plessis_class_2014}. Positive-unlabelled can also be framed as a cost-sensitive learning problem \cite{du_plessis_analysis_2014}. Under some conditions it has been showed that positive-unlabelled learning can outperform positive-negative learning if the unlabelled set is large enough \cite{niu_theoretical_2016}. Multi-positive-unlabelled learning is a generalization of positive-unlabelled learning, in which there are multiple positive classes. There has been limited work in this area, but \cite{xu_multi-positive_2017} proposes a generic loss-function and optimizes a linear classifier for this problem. \\

In noisy-label learning we explicitly assume that some training labels may be incorrect (noisy). Learning the underlying concepts thus requires robust models like Robust Multi-class Gaussian Process Classifiers \cite{hernandez-lobato_robust_2011} or robust deep neural network \cite{ma_dimensionality-driven_2018}. A transductive approach is to relabel training samples according to the belief of the available model(s). Ensemble methods can be used to predict probabilities of being correct and identify noisy samples \cite{brodley_identifying_1999}. Deep models can also be used to create parallel correction systems for correcting labels \cite{han_deep_2019}. Most work considers constant noise, but some handle class-conditional noise \cite{natarajan_learning_2013, xiao_learning_2015, sukhbaatar_learning_2014}. DivideMix \cite{li_dividemix_2020} combines MixMatch and co-dividing (two models are trained to predict labels for each other) to transfer the methodologies from semi-supervised learning to noisy-label learning. 

\section{Decoupling}

Consider a dataset with $n$ samples. Each sample has one label and these labels are gathered in a one-hot encoded matrix
\begin{align*}
	&\mS_{\mathcal{D}} \in \{0, 1\}^{n \times m_s}, \quad \mS_{\mathcal{D}} \vone = \vone,
\end{align*}
where $m_s$ is the number of possible selection labels and the $\vone$'s are vectors of ones (of suitable dimensionality). All samples have exactly one label and we will therefore have a dedicated label for "unlabelled samples" if needed. 

Using machine learning techniques we can train a model to predict the \textit{label}-distribution $p(s \given \vx)$. If this model is good we can use it to approximate the following matrix containing the \textit{selection} probabilities of the samples (or new samples)
\begin{align}
	&\mS \in [0, 1]^{n \times m_s}, 
		\quad 0 \leq \mS_{is} = p(s \given \vx_i), 
		\quad \mS \vone = \vone.
\end{align}

While the selection probabilities may relate somewhat to the classes, they are not what we are searching for. We wish to determine the \textit{class} probabilities, which will here gather in the (unknown) matrix
\begin{align}
	&\mY \in [0, 1]^{n \times m_y}, 
		\quad 0 \leq \mY_{i y} = p(y \given \vx_i), 
		\quad \mY \vone = \vone,
\end{align}
where $m_y$ is the number of classes.

We assume random sampling of selection labels within the classes, so $\vx$ and $s$ are conditionally independent given $y$:
\begin{align*}
	p(s \given y, \vx) = p(s \given y).
\end{align*}
Note that in general the opposite assumption does not hold: $p(y \given s, \vx) \not= p(y \given s)$. 
The probability of a selection $s$ for a sample becomes
\begin{align}
	p(s \given \vx) &= \sum_{y} p(s \given y, \vx) \; p(y \given \vx) \nonumber = \sum_{y} p(s \given y) \; p(y \given \vx). \label{eq:selection_from_classes}
\end{align}
For a set of samples this can be expressed as a set of linear equations by
\begin{align}
	\mS = \mY\mT,
\end{align}

where we call $\mT$ the \textit{transition} matrix, as is customary in noisy-label learning and for similar variables in for example Markov processes. $\mT$ is defined by
\begin{align*}
	&\mT \in [0, 1]^{m_y \times m_s}, 
	\quad 0 \leq \mT_{y s} = p(s \given y), 
	\quad \mT \vone = \vone.
\end{align*}

We can now define the \textit{decoupling problem} as
\begin{center}
	\begin{minipage}{0.9\linewidth}
	\begin{center}
		\textit{Inferring the class distributions $p(y \given \vx)$ from the transition
		distributions $p(s \given y)$ and the label distributions $p(s \given \vx)$.}
	\end{center}
	\end{minipage}
\end{center}

That is, we want to infer $\mY$ from our $\mS$ and $\mT$, while considering any uncertainty about $\mS$ and $\mT$. \\

While it may be tempting to isolate $\mY$ using the inverse or pseudo-inverse of $\mT$, for most situations this is not a suitable approach and will usually results in negative and unscaled values (for the probabilities). Furthermore we may not know $\mT$ exactly. Instead we use a variational inference method for optimizing our belief about $\mY$ and $\mT$. \\

In the classical positive-unlabelled case, the probability for the negative class being labelled as positive is zero. Thus $\mT$ is square and has three non-zero values. If one applies the inverse of $\mT$ we arrive at the method of Elkan and Noto \cite{elkan_learning_2008}, which we have shown in supplementary section \ref{append:positive_unlabelled_learning_elkan}, \textit{\nameref{append:positive_unlabelled_learning_elkan}}. The decoupling problem is therefore a generalization of the positive unlabelled learning problem in earlier works.

\section{Generative Process}

\subsection{Assumed Distributions}
In order to work with the decoupling problem, we assume useful distributions on each of the involved variables. \\

We assume the transition probabilities follow a Dirichlet distribution for each class/row. 
\begin{align}
	p(\mT) &= \prod_{y} \dirichlet_y(\mT_{y:}) = \frac{1}{B(\mA)} \prod_{ys} \emT_{ys}^{\emA_{ys} - 1}, \label{eq:T_prior} 
\end{align}
using (\ref{eq:def_multivar_beta_function_matrix}) for normalization. \\
We similarly assume a Dirichlet distribution for each sample's class-distribution
\begin{align}
	p(\mY \given \vgamma) = \frac{1}{B(\vgamma)} \prod_{iy} \emY_{i y}^{\evgamma_y - 1}, \label{eq:Y_prior}
\end{align}
with parameters $\vgamma$. \\
We may have some uncertainty on $\vgamma$ and thus employ a conjugate prior of the Dirichlet distribution \cite{andreoli_conjugate_2018}
\begin{align}
	p( \vgamma ) = \frac{1}{Z(\eta, \vlambda) B(\vlambda)^{\eta}} \e^{- \sum_y \evlambda_y \evgamma_y}, \label{eq:Y_prior_distribution}
\end{align}
where $\eta$ and $\vlambda$ are parameters.\\
The probability of a selection matrix conditioned on classes and transitions is
\begin{align}
	p(\mS \given \mY, \mT) = \prod_{is} (\mY \mT)_{is}^{\emS_{is}}, \label{eq:S_likelihood}
\end{align}
and becomes the likelihood of our problem. \\
The joint distribution takes the form
\begin{align}
	p(\mS&, \mY, \mT, \vgamma) 
		= p(\mS \given \mY, \mT) \: p(\mY \given \vgamma) \: p(\mT) \: p(\vgamma)
		\label{eq:problem_joint_distribution} \\
		%
		%
		&= 
			\frac{1}{
				B(\vgamma)
				B(\mA)
				Z(\eta, \vlambda) B(\vlambda)^{\eta}
			} \times \nonumber \\
		&\qquad\qquad
			\prod_{iys} (\mY \mT)_{is}^{\emS_{is}}
			\emY_{i y}^{\evgamma_y - 1}
			\emT_{ys}^{\emA_{ys} - 1}
			\e^{-\evlambda_y \evgamma_y}.
		\nonumber
\end{align}

\begin{figure}
	\centering
	\includegraphics[width=0.8\linewidth, valign=t]{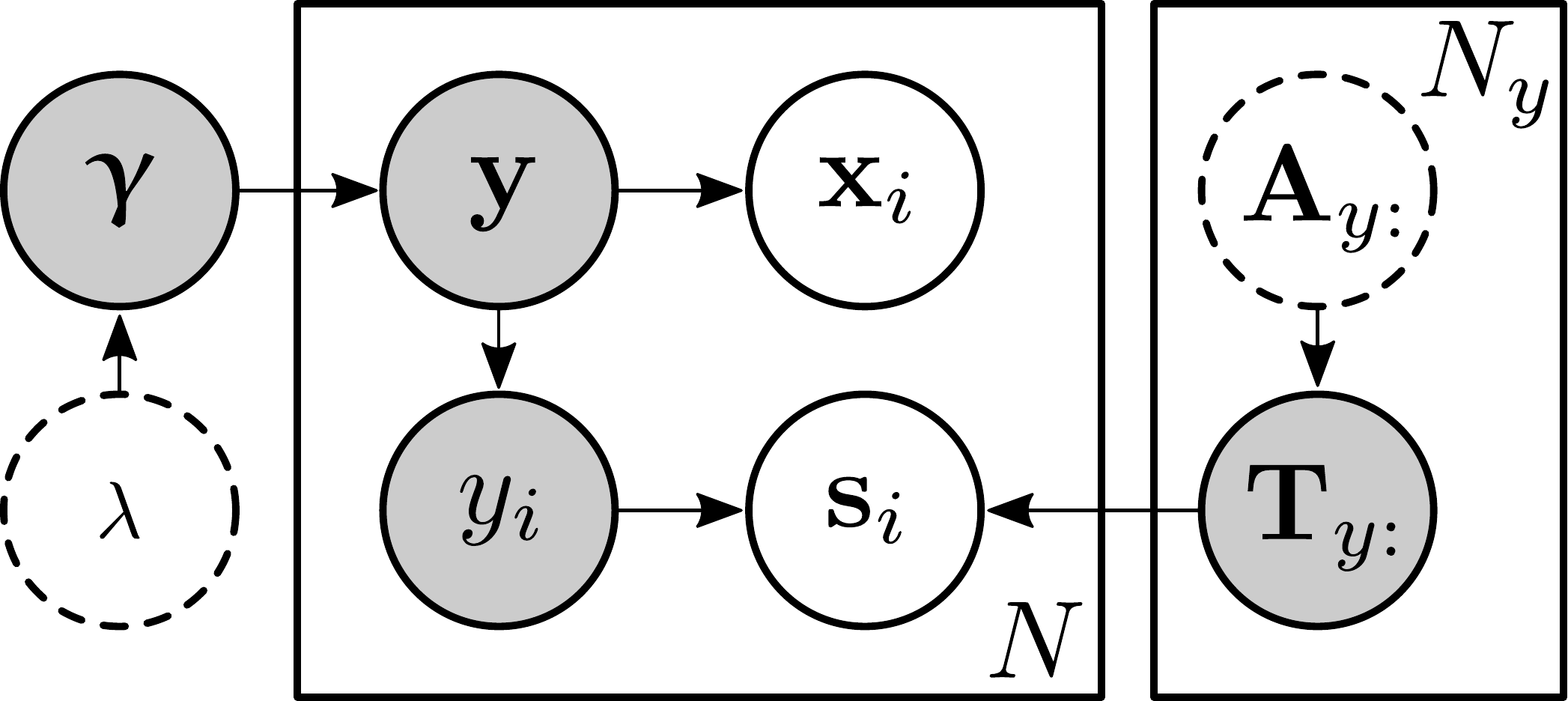}
	\caption{Graphical model of sampling process. The white, solid nodes are observations, the nodes with dashed lines are priors, and the rest are unobserved.}\label{fig:graphical_plate}
\end{figure}

\subsection{Process}
The following generative processes describes the assumptions on the system and is illustrated by the graphical model in figure \ref{fig:graphical_plate}.
\begin{enumerate}
	\item Draw a class prior: $\vgamma \sim \Dirp(\lambda, \eta)$.
	\item Draw transition distributions for classes: \\$\mT_{y:} \sim \dirichlet(\mA_{y:})$.
	\item For each sample $i$
		\begin{enumerate}
		\item Draw a class distribution: $\vy \sim \dirichlet(\vgamma)$.  \label{enum:generative_draw_class_distribution}
		\item Draw a class: $y_i \sim \categorical(\vy)$.  \label{enum:generative_draw_class}
		\item Draw $\vx_i$ from some underlying generative \\ distribution $p(\vx \given \vy)$.
		\item Draw a selection distribution: $\vs_i \sim \dirichlet(\mT_{y_i:})$.
		\end{enumerate}
\end{enumerate}~\\[-1cm]

%

\subsection{Prior on Class Distributions}

In the generative process before, consider the (prior) probability of each class 
\begin{align}
	p(y \given \vgamma) &= \int 
		\evpi_y
	 	\cdot 
	 	\frac{1}{B(\vgamma)} \prod_{ij} \evpi_j^{\evgamma_j - 1}
	 	\d \vpi \\
 		&= \E_{\dirichlet, \vgamma}[\evpi_y], \nonumber
\end{align}
where $\evpi_y = p(y)$. This is the expectation under the Dirichlet distribution given by the parameters $\vgamma$. Thus $\vgamma$ acts like a prior on classes, but also controls the shape of the Dirichlet distribution on classes. The Dirichlet distribution parameterized by $\vgamma$ will have most of its mass at the edge-points (where most probability is for a single class) if $\sum_i \evgamma_i < 1$, and will have most of its mass towards its expectation if $\sum_i \evgamma_i > 1$. The parameters in $\vgamma$ therefore also control how much the classes are expected to overlap. Choosing small values for $\vgamma$ will make the class-distributions more sparse, and put more mass on the most probable classes. Choosing large values for $\vgamma$ will make the system more conservative and keep a bit of probability for all classes for most samples.

\section{Variational Distributions}

We create a variational distribution for $\mY$ and $\mT$, with parameters to be learned. We choose a Dirichlet distribution on each row of $\mT$ and on each row of $\mY$. We use the respective parameters $\mTheta$ and $\mPi$ of appropriate sizes. We discuss variational distributions for $\vgamma$ later.
\begin{align}
	q(\mT_{y:}) &= \frac{1}{B(\mTheta)} \prod_s \emT_{ys}^{\emTheta_{ys} - 1}, \\
	q(\mY_{i:}) &= \frac{1}{B(\mPi)} \prod_y \emY_{iy}^{\emPi_{iy} - 1}.  \label{eq:variational_approximations_q}
\end{align}

\subsection{Optimal Model}

The optimal models are ideally determined by the Kullback-Leibler divergence
\begin{align}
	&\amax_{\mTheta, \mPi, p(\vgamma)} \text{KL}_p[q(\mS, \mY, \mT, \vgamma), p(\mS, \mY, \mT, \vgamma)],
\end{align}
which is unfortunately not tractable with these distributions using known methods.

\subsection{Evidence Lower Bound (ELBO)}


We optimize our model by maximizing the Evidence Lower BOund (ELBO), which for our problem is (detailed in supplementary material \ref{appendix:elbo_derivation}, \textit{\nameref{appendix:elbo_derivation}})
\begin{flalign}
	\text{ELBO} 
		&= \E_q[ \ln p(\mS \given \mY, \mT) ] 
			+ \E_q[ \ln p(\mY \given \vgamma) \: p(\mT) \: p(\vgamma) ] \nonumber \\
			&- \E_q[ \ln q(\mY) \: q(\mT) \: q(\vgamma) ] 
\end{flalign}\vspace{-5mm}
\begin{align}
		&= \sum_{is} \emS_{is} \int_{q\triangle} \ln \Big( \sum_{y} \emY_{iy} \emT_{ys} \Big) \d \mT \d \mY
			 &&{\color{sRed}\bm{\bigg\}}} \: \text{(L)} \nonumber \\
			&\qquad +
				\sum_{iy} \big(\E_q[\evgamma_y] - 1\big) \E_q\big[ \ln \emY_{i y} \big]
				 &&{\color{s_sek_blue}\bm{\bigg\}}} \: \text{(P)} \nonumber \\
			&\qquad + \sum_{ys} (\emA_{ys} - 1) \E_q\big[ \ln \emT_{ys} \big]
				- \sum_y \evlambda_y \E_q\big[ \evgamma_y \big] 
				 \hspace{-3.5mm}&&{\color{s_sek_blue}\bm{\bigg\}}} \: \text{(P)} \nonumber \\
			&\qquad 
			    - \sum_i H_{\dirichlet}(\mY_{i:})
				- \sum_y H_{\dirichlet}(\mT_{y:})
				 &&{\color{s_sek_blue}\bm{\bigg\}}} \: \text{(E)} \nonumber \\
			&\qquad
				- H_{q}(\vgamma) - \E_q[\ln B(\vgamma)] 
				 &&{\color{sRed}\bm{\bigg\}}} \: \text{(Cl)} \nonumber \\
			&\qquad - \ln B(\mA) - \ln \Big( Z(\eta, \vlambda) B(\vlambda)^{\eta} \Big),
			 &&{\color{s_sek_green}\bm{\bigg\}}} \: \text{(Co)} \nonumber
\end{align}
\begin{description}
	\item[(L)] Likelihood which is problematic
	\item[(P)] Priors with analytic solutions
	\item[(E)] Entropies with analytic solutions
	\item[(Cl)] Class prior, unknown
	\item[(Co)] Constants
\end{description}

There is a few things to note about the ELBO. First of all the last line contains constants and can be disregarded for optimization. Secondly the likelihood is problematic to compute, but we show a method for approximating this quantity in the following section. \\

Finally (Cl) contains elements regarding the distribution of the class prior ($\vgamma$ also appear in one of the prior terms). This line shows a peculiar property which we need for the variational distribution on $\vgamma$; we need to be able to compute the expectation of the logarithm of the multivariate binomial coefficient of the elements. For now we do not suggest a variational distribution on $\vgamma$, but we do note that in future work this quantity can be sampled, as it only scales linearly in the number of classes $O(m_y)$. For the tests in this paper we instead assume $\vgamma$ to be known (ELBO in this case can be seen in supplementary material \ref{appendix:elbo_with_constant_gamma}, \textit{\nameref{appendix:elbo_with_constant_gamma}}).

\section{Approximation Methods for Decoupling} \label{sec:approximation_methods_decoupling}

The ELBO is difficult to compute due to the likelihood-term, which has an expectation over a logarithm of a sum of stochastic variables. We have found one way to approximate this quantity. Our approximation scheme is quite lengthy and most of the details  are found in the supplementary material. \\
The main points are
\begin{enumerate}[label={\large \textbf{\Roman*.~~~}}, itemsep=3ex]
	\item One approximation of the expectation of the natural logarithm, based on the Taylor expansion around the mean, is
	\begin{align}
		\E[\ln(X)]
			= \ln(\mu) + \sum_{p=1}^K (-1)^{p - 1} \frac{1}{p \cdot \mu^p} \: \mu_p,
	\end{align}
	where $\mu_p=\E\big[ ( X - \E[X] )^n \big]$ is the $p$'th moment of $X$ and $\mu$ is the expectation.
	The details are in supplementary material \ref{appendix:expected_log_approximation}, \textit{\nameref{appendix:expected_log_approximation}}.
	\item The above approximation will be applied to a sum of variables. Each approximation-term will be thus expanded to a sum of many products of variables, due to the exponent of the moments. 
	We define the following function 
	\begin{align}
		\xi(\vx, p) = \bigg(\sum_k^K \evx_k\bigg)^p
	\end{align}
	which we analyse and conclude scales by not quite $K^p$ but remains exponential (it scales by the number of term combinations as outlined in supplementary material \ref{appendix:power_of_sum}, \textit{\nameref{appendix:power_of_sum}}). For our problem it is possible to compute all these quantities.
	
%
	
	%
	%
	%
	%
	\item We define the following tensor
	\begin{align}
		&\tS \in \R^{n \times m_s \times m_y}, \qquad
		\etS_{isy} = \emY_{iy} \emT_{ys}, \\
		&\sum_{y} \emY_{iy} \emT_{ys} = \vone^{\top} \tS_{is:}. \nonumber
	\end{align}
	
	We can express the expected value of the exponentiated sum over $\mY$-$\mT$-products as
	\begin{align}
		\E&\Bigg[ \bigg( \sum_{y} \emY_{iy} \emT_{ys} \bigg)^p \Bigg]
			= \E\Big[ \xi\big(\tS_{is:}, p\big) \Big] \\
			&= \!\!\!\sum_{ \vbeta \in \mathcal{B}^{(K, n)} } \!\!\! 
				{n \choose \vbeta} 
				\E_q\bigg[ \prod_y \emY_{iy}^{\vbeta_y} \bigg]
				\E_q\bigg[ \prod_y \emT_{ys}^{\vbeta_y} \bigg]. \nonumber
	\end{align}
	We refer to section \ref{appendix:power_of_sum}, \textit{\nameref{appendix:power_of_sum}} for details on $\xi(\cdot)$ and the set $\mathcal{B}^{(K, n)}$. 
	
	Using the above approximation and notations we can express an approximation of the expected logarithm of the sum over $\mY$-$\mT$-products as
	\begin{align}
		&\E \Big[ \ln \Big( \sum_{y} \emY_{iy} \emT_{ys} \Big) \Big]  \\
			&\approx \ln\Big( \sum_{y} \E\big[ \emY_{iy} \emT_{ys} \big] \Big) - \sum_{p=1}^K  \frac{1}{p} \nonumber \\
				&\:\:\:
				+ \sum_{j=1}^K \tau_j 
					\bigg( \sum_{y} \E\Big[ \emY_{iy} \emT_{ys} \Big] \bigg)^{-j} 
					\E\Big[ \xi\big(\tS_{is:}, j\big) \Big], \nonumber
	\end{align} 
	where the coefficients are
	\begin{align}
		\tau_j &= (-1)^{j - 1} \sum_{p=1}^K  \frac{1}{p} {p \choose j}.
	\end{align}
	
	The derivations of the above are found in supplementary material \ref{appendix:expected_log_approximation_for_elbo_likelihood}, \textit{\nameref{appendix:expected_log_approximation_for_elbo_likelihood}}. \\
	
	We call $\E_q\big[ \xi(\tS_{is:}, n) \big]$ the \textit{agreement term of order $n$}. The agreement terms represents how well each sample's class-distribution "agrees" with a given selection. We investigate this and show some examples in supplementary material \ref{appendix:agreement_terms}, \textit{\nameref{appendix:agreement_terms}}. Furthermore we can straightforward compute the derivatives of these approximations for optimization. 
	\item We can now approximate the ELBO likelihood in time complexity $O(n \times m_s \times m_y^K)$, where $K$ is the order of the approximating Taylor series.
	\item We furthermore arrange and approximate the label-conditional class probabilities in the following matrix. Details can be found in \ref{appendix:label_conditional_class_probabilities}, \textit{\nameref{appendix:label_conditional_class_probabilities}}.
	\begin{gather}
		\mW \in [0, 1]^{n \times m_y}, 
			\quad \mW \vone = \vone, \\[2mm]
		0 \leq \mW_{i y} = p(y \given s_i, \vx_i) \approx \frac{
				\E_q[\emT_{ys_i}] \E_q[\emY_{iy}]
			}{
				\sum_{y'} \E_q[\emT_{y's_i}] \E_q[\emY_{iy'}]
			}. \nonumber
	\end{gather}
\end{enumerate}

\section{Simulated Example}

\begin{figure}[h!b]
	\centering
	\includegraphics
	[width=0.7\linewidth, page=2]{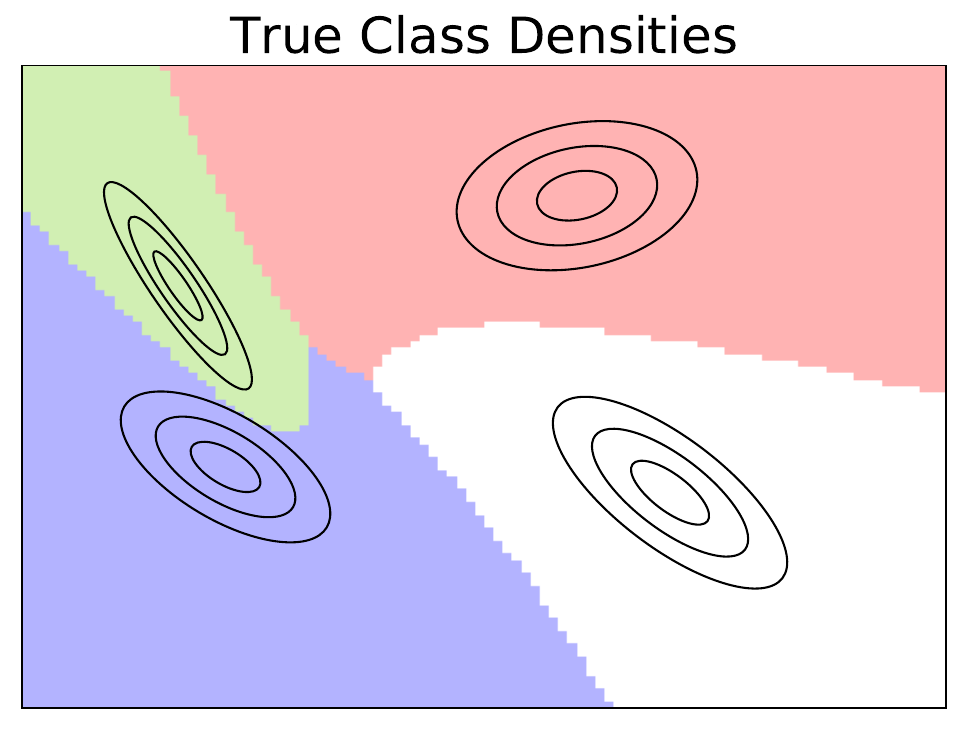}
	\caption{True distributions are four normals, here shown with decision boundaries. White is negative class.} \label{fig:simulated_normals}
\end{figure}

\begin{figure}[h!b]
	\centering
	\includegraphics
	[width=0.9\linewidth, page=4]{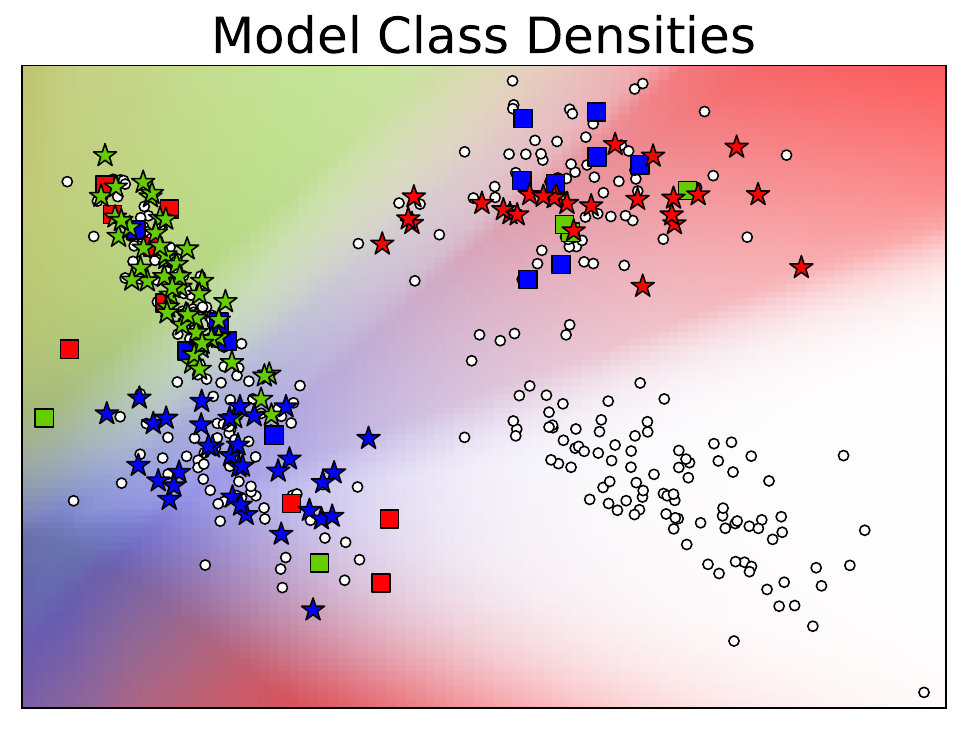}
	\caption{Sampled dataset with class densities from trained neural network. Stars are correctly labelled, squares are noisy (incorrect) labels and circles are unlabelled.} \label{fig:simulated_model}
\end{figure}

\begin{figure}[h!]
	\centering
	\includegraphics
	[width=0.9\linewidth, page=5]{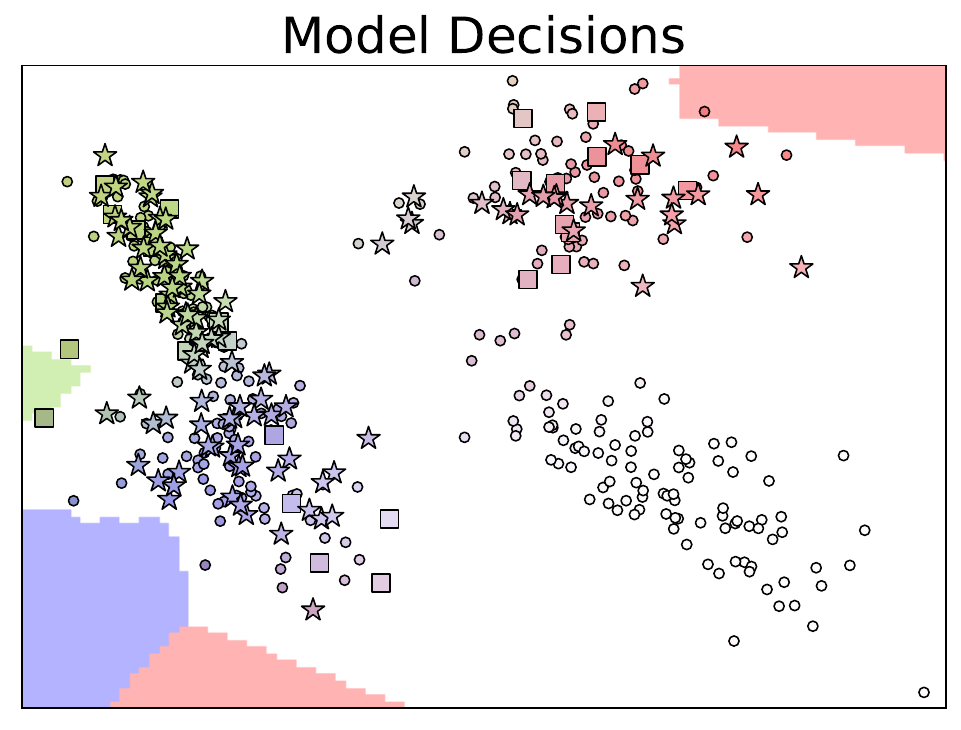}
	\caption{Decision boundaries and predictions on samples by the same neural network. Most regions are classified as "unlabelled", which is the most probable label.} \label{fig:simulated_model_decisions}
\end{figure}

\begin{figure}[h!]
	\centering
	\includegraphics
	[width=0.9\linewidth, page=6]{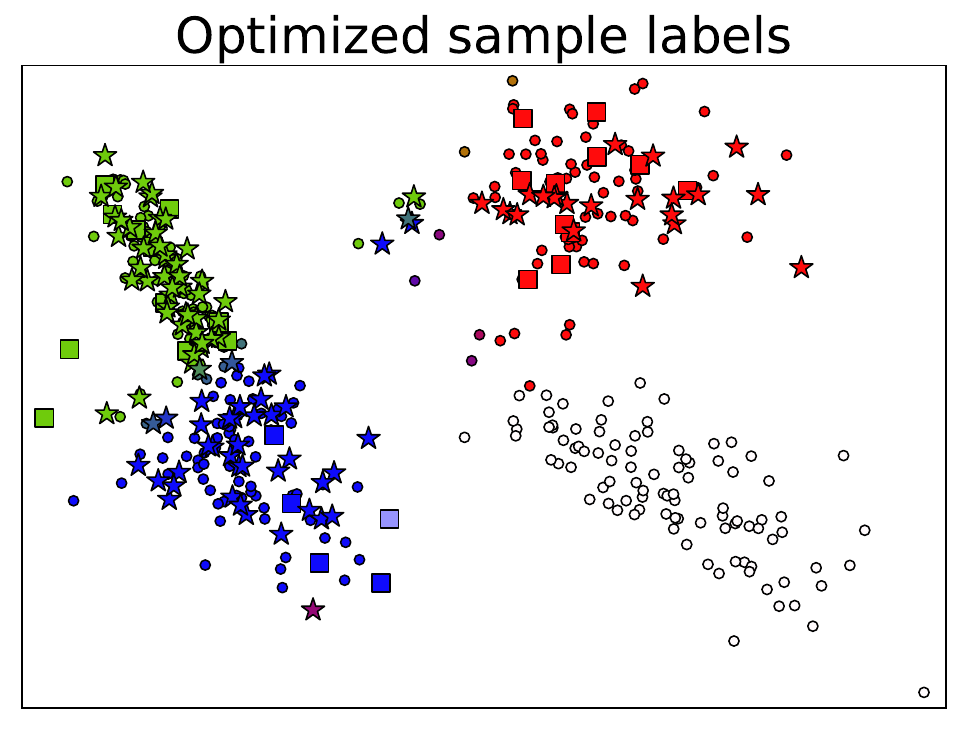}
	\caption{The optimized class-probabilities of samples. Many samples are correctly labelled and "mixed" regions have uncertainty in their labels.}
	\label{fig:simulated_optimized}
\end{figure}

\begin{figure}[h!]
	\centering
	\includegraphics
	[width=\linewidth, page=10]{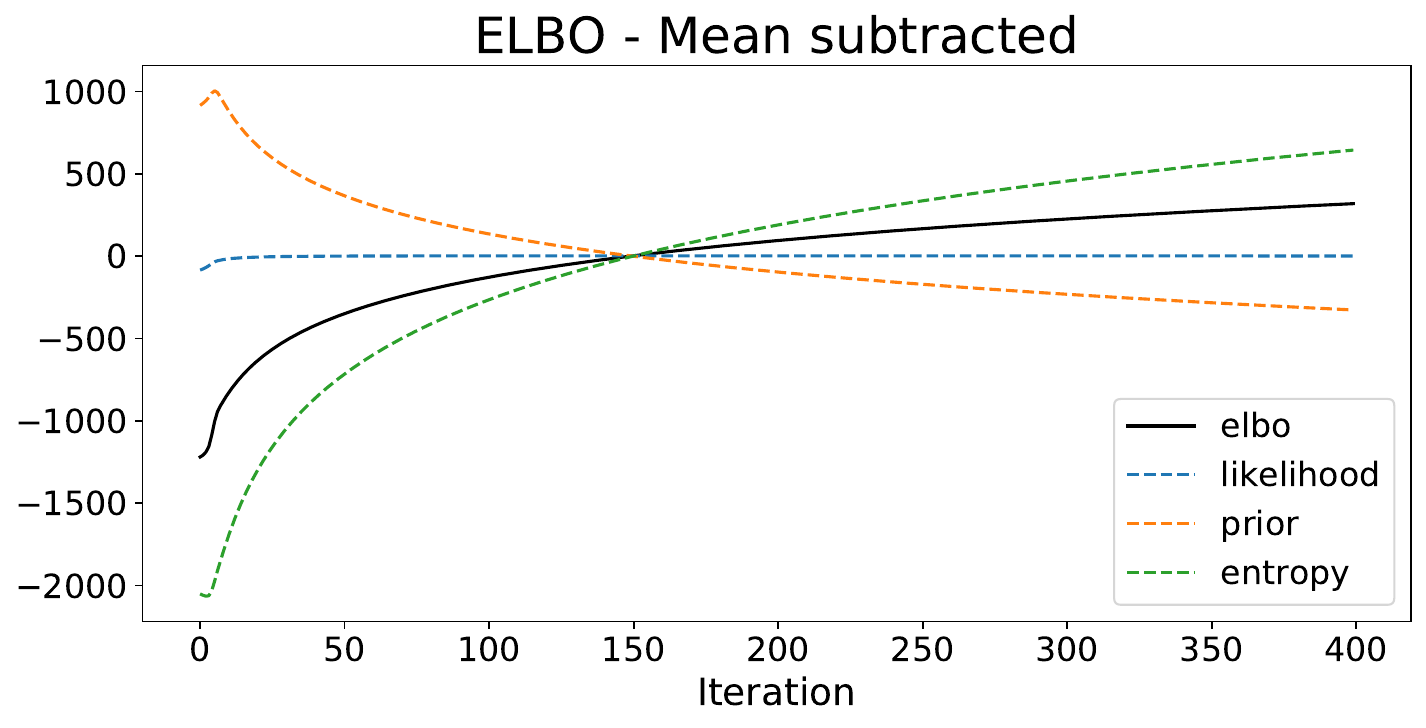}
	\caption{ELBO components during optimization (mean of each curve is subtracted for readability). The prior-component is decreased while the entropy-component is increased.}
	\label{fig:simulated_elbo_curve}
\end{figure}

\begin{figure}[h!]
	\centering
	\includegraphics
	[width=\linewidth, page=11]{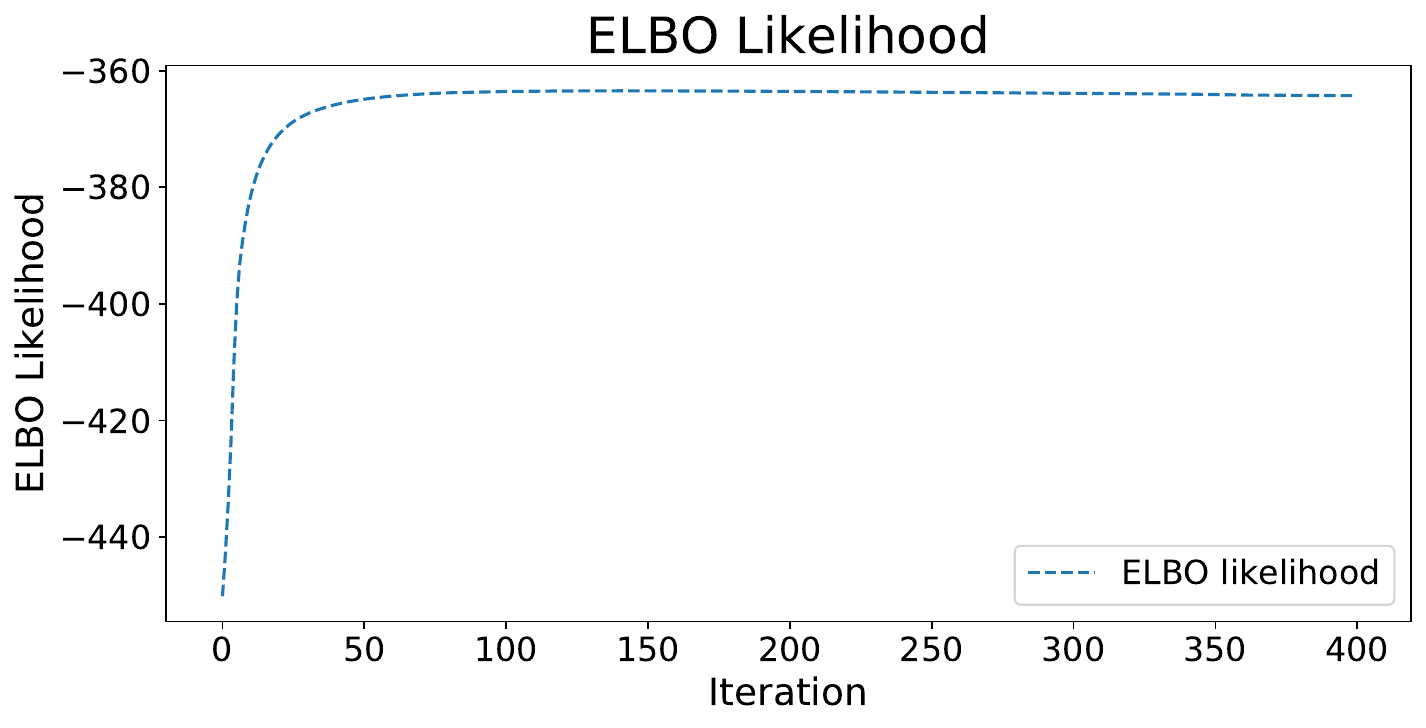}
	\caption{A zoom of the likelihood component. This component is first increased and then flattens out. The likelihood will sometimes decrease to give way to the entropy component. } \label{fig:simulated_likelihood_curve}
\end{figure}

Figures \ref{fig:simulated_normals}-\ref{fig:simulated_likelihood_curve} show how inferring class distributions for samples works on simulated data. We simulate a multi-positive-unlabelled, noisy-label classification problem, with unlabelled probability $p(\text{unlabel}) = 0.7$, noise between positive labels $p(\text{noise}) = 0.05$ and noise between negative label and positive labels $p(\text{neg-noise}) = 0.0025$. The figure-captions outline the experiment.

\newpage
\section{Decoupling on Real Dataset}

For testing out the inference of class probabilities, we use the Fashion MNIST dataset\footnote{~A commonly used benchmark dataset, containing 60.000 training set samples and 10.000 test samples. Each sample is a 28x28 grayscale image, with an associated label from 10 classes. The label represents a type of clothing depicted in the image.} \cite{xiao_fashion-mnist_2017}.
We use the dataset to simulate 6 tasks, and provide labelled and unlabelled samples for each task (all tasks are semi-supervised). The tasks are; semi-supervised (use dataset as is), positive-unlabelled (class 1 is positive, the rest are negative), 7-positive-unlabelled (classes 1-3 are positive, the rest are negative), noisy-label learning with noise-rate 20\% and 50\%, and class-conditional noisy-label learning. In the class-conditional noisy-label learning we make all classes have a 22\% chance of having their label flipped to a class with lower class number (class 3 can be flipped to classes 1 or 2 etc.). Since the first class has no noise the average noise-rate becomes around 20\%. 
We vary the number of per-class labelled samples for each task to be 100, 200, 400, 800, 1.000, 1.600, 2.400, 3.200 and 5950 (there are 6.000 samples per dataset class). 
We use the F1-score to measure performance (Section \ref{append:f_scores}, \textit{\nameref{append:f_scores}}), as it is useful for unbalanced datasets with varying number of classes, making the tasks comparable. \\

For each task we train a simple neural network (1 convolutional layer with pooling followed by a single dense layer, trained with a bit of regularization) to predict the \textit{label} distribution. As we are attempting to infer class-probabilities on the training data itself, it is crucial to use a well-regularized model which can not easily overfit the training data. This motivates the use of this fairly simple neural network.
We optimize the previously mentioned ELBO in order to infer the \textit{class} distributions from the predicted \textit{label} distributions. The plots are shown with three standard deviations based on 10 runs, although the curves are very stable except for when there are very few samples. 

%

\renewcommand{\temp}{0.25}
\begin{figure}[h!]
	\centering
	\begin{subfigure}{\linewidth}
		\centering
		\includegraphics[width=0.5\linewidth,page=14, valign=m]{fig/python/neural_results_massive.pdf}
	\end{subfigure} \\[2mm]
	\begin{subfigure}{\linewidth}
		\centering
		\includegraphics[width=\linewidth,page=1, valign=m]{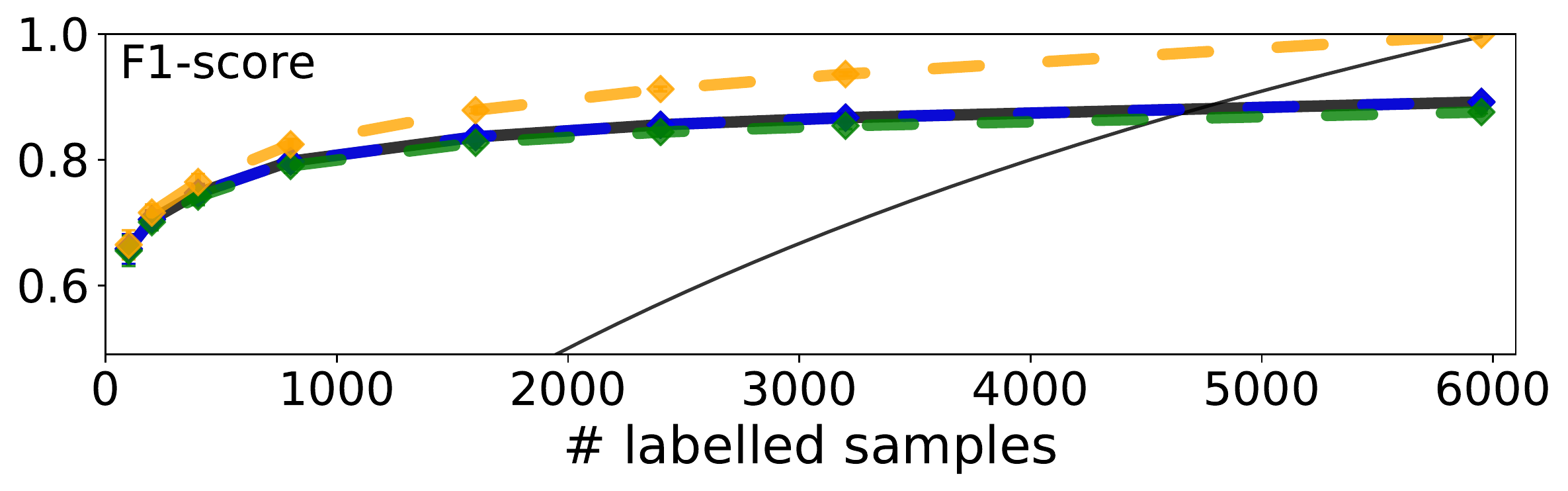}
		\caption{Semi-supervised.}  \label{subfig:fashion_semisupervised}
	\end{subfigure} \\[2mm]
	\begin{subfigure}{\linewidth}
		\centering
		\includegraphics[width=\linewidth,page=4, valign=m]{fig/python/neural_results_massive.pdf}
		\caption{Noisy-labelling - 20\%.}  \label{subfig:fashion_noisy_low}
	\end{subfigure} \\[2mm]
	\begin{subfigure}{\linewidth}
		\centering
		\includegraphics[width=\linewidth,page=5, valign=m]{fig/python/neural_results_massive.pdf}
		\caption{Noisy-labelling - 50\%.}  \label{subfig:fashion_noisy_high}
	\end{subfigure} \\[2mm]
	\begin{subfigure}[t]{\linewidth}
		\centering
		\includegraphics[width=\linewidth,page=6, valign=m]{fig/python/neural_results_massive.pdf}
		\caption{Noisy-labelling - conditional 20\%.}  \label{subfig:fashion_conditional_noise}
	\end{subfigure}
	\begin{subfigure}[t]{\linewidth}
		\centering
		\includegraphics[width=\linewidth,page=3, valign=m]{fig/python/neural_results_massive.pdf}
		\caption{Positive-unlabelled.}  \label{subfig:fashion_pu}
	\end{subfigure} \\[2mm]
	\begin{subfigure}[t]{\linewidth}
		\centering
		\includegraphics[width=\linewidth,page=2, valign=m]{fig/python/neural_results_massive.pdf}
		\caption{7-positive-unlabelled.}  \label{subfig:fashion_7_positive}
	\end{subfigure}
	\caption{Performance on Fashion MNIST dataset. Notice that the y-axis of the positive-unlabelled and 7-positive-unlabelled graphs are from 0.0 to 1.0, while the rest are from 0.5 to 1.0.}  \label{fig:fashion}
\end{figure}

Figure \ref{subfig:fashion_semisupervised} shows how the training performance of semi-supervised task follows the performance of the label-classifier, which we can show will generally be the case for this simple problem (see supplementary section \ref{append:super_and_semisuper_learning}, \textit{\nameref{append:super_and_semisuper_learning}} for details about how decoupling handles semi-supervised learning). The test performance is a bit lower as it tends to happen with machine learning models. More interestingly we see that the label-conditional performance can be used to impute labels to the unlabelled set, in a well-defined way. We do not tell the model to only impute labels on the unlabelled data, but due to the uncertainty in the transitions of this label the model does this automatically. For noisy-label learning we see a similar phenomenon with 20\% noise rate (figure \ref{subfig:fashion_noisy_low}), although the label-conditional performance is a bit lower due to the noise on those labels. Again we can use the label-conditional method to impute labels on unlabelled data and even correct noisy labels. With 50\% noise (figure \ref{subfig:fashion_noisy_high}), the labels are so bad that they no longer help and the performance simply becomes that of the underlying model. In the class-conditional noise case (figure \ref{subfig:fashion_conditional_noise}) we see a result similar to that of 20\% noise rate. We are able to encode the noise-structure and impute some labels. \\
For positive-unlabelled and multi-positive-unlabelled learning the inference system is able to heavily improve performance of the model for both training and test data. For both situations there is a region where using the labels can further improve performance, but for the 7-multi-positive case, using labels decreases performance around 2000 labels, likely due to small confidence in positive labels.

\section{Self-Training}

SETRED \cite{li_setred_2005} is a self-training approach which combines a classifier with graph-based methods for predicting labels on unlabelled data. We test using this model in the positive-unlabelled and multi-positive-unlabelled setting. It was not made for these problems, but we hypothesise that decoupling can help the model. For computing edges for the graph in SETRED we use the internal representation of the images, by the classifying neural network. 
The reasoning for using SETRED is that we can somewhat easily modify it into other non-standard classification tasks than semi-supervised learning, and because it scored as one of the best performing systems in the survey by Triguero et al. \cite{triguero_self-labeled_2015}. Another self-training method that also scored high in that survey is on by Wang et al. \cite{wang_semi-supervised_2010}, but this method relies directly on a distance-measure between samples, which can be tricky for image data. \\

In the following we use the term \textit{relabelled samples} to refer to unlabelled samples that have been given a label by the model. We are here only concerned with the performance on the relabelled samples to reflect the transductive performance of the systems. For all experiments we use 1000 labelled samples from each class, expect for positive-unlabelled learning where we use 3000 (this problem was a bit harder), and use the same neural network as the previous experiments. \\

\begin{figure}[h!]
	\centering
	\begin{subfigure}{\linewidth}
		\centering
		\includegraphics[width=0.7\linewidth,page=6, valign=m]{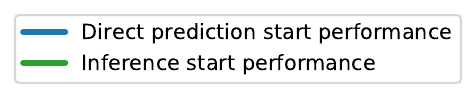}
	\end{subfigure} \\[2mm]
	\begin{subfigure}{\linewidth}
		\centering
		\includegraphics[width=\linewidth,page=2, valign=m]{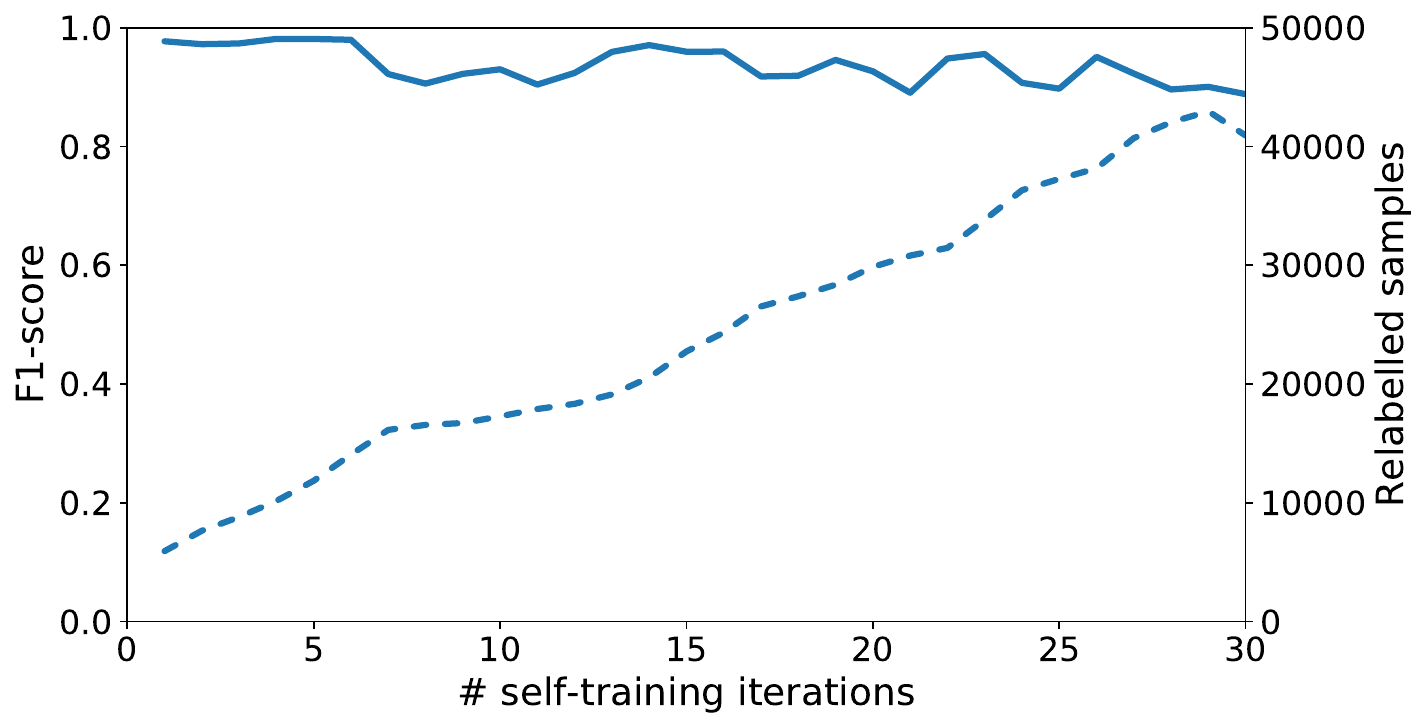}
		\caption{Semi-supervised.}  \label{subfig:self_training_semisupervised}
	\end{subfigure} \\[2mm]
	\begin{subfigure}{\linewidth}
		\centering
		\includegraphics[width=\linewidth,page=3, valign=m]{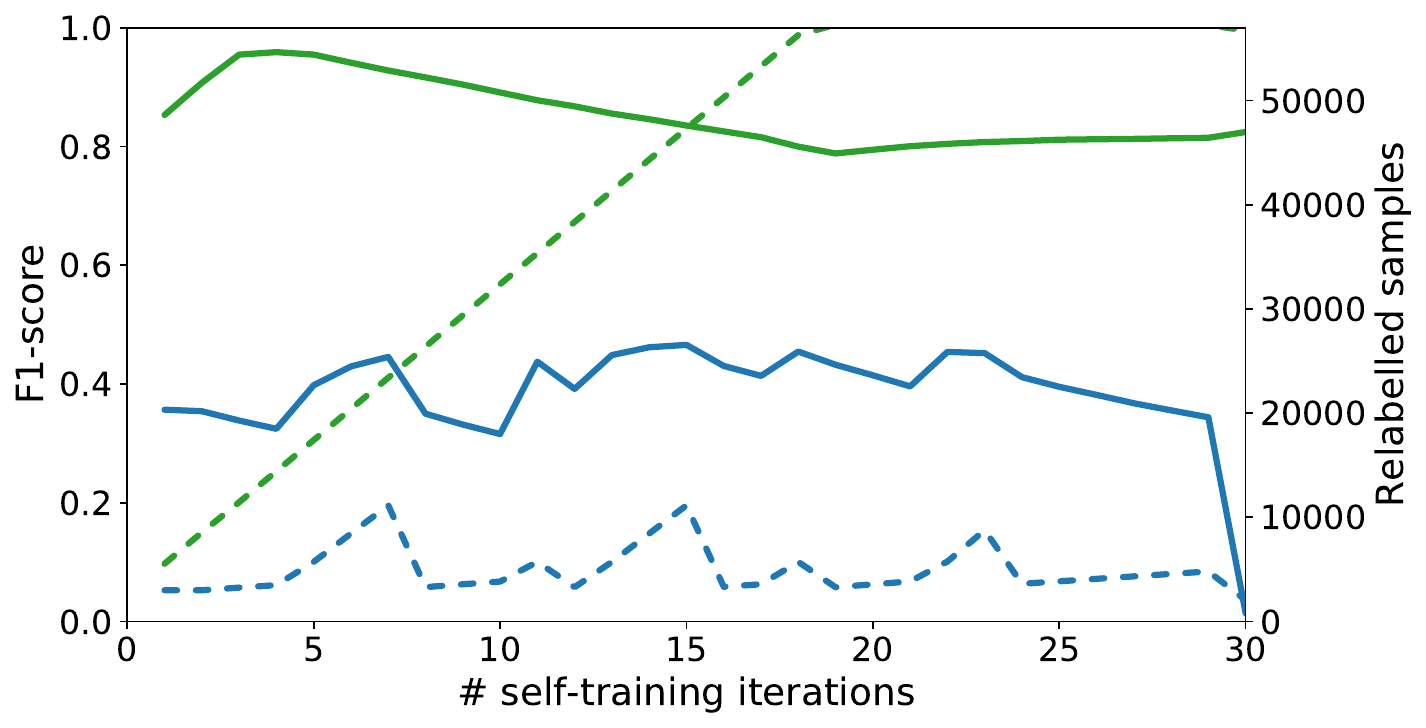}
		\caption{Positive-unlabelled.}  \label{subfig:self_training_pu}
	\end{subfigure} \\[2mm]
	\begin{subfigure}{\linewidth}
		\centering
		\includegraphics[width=\linewidth,page=4, valign=m]{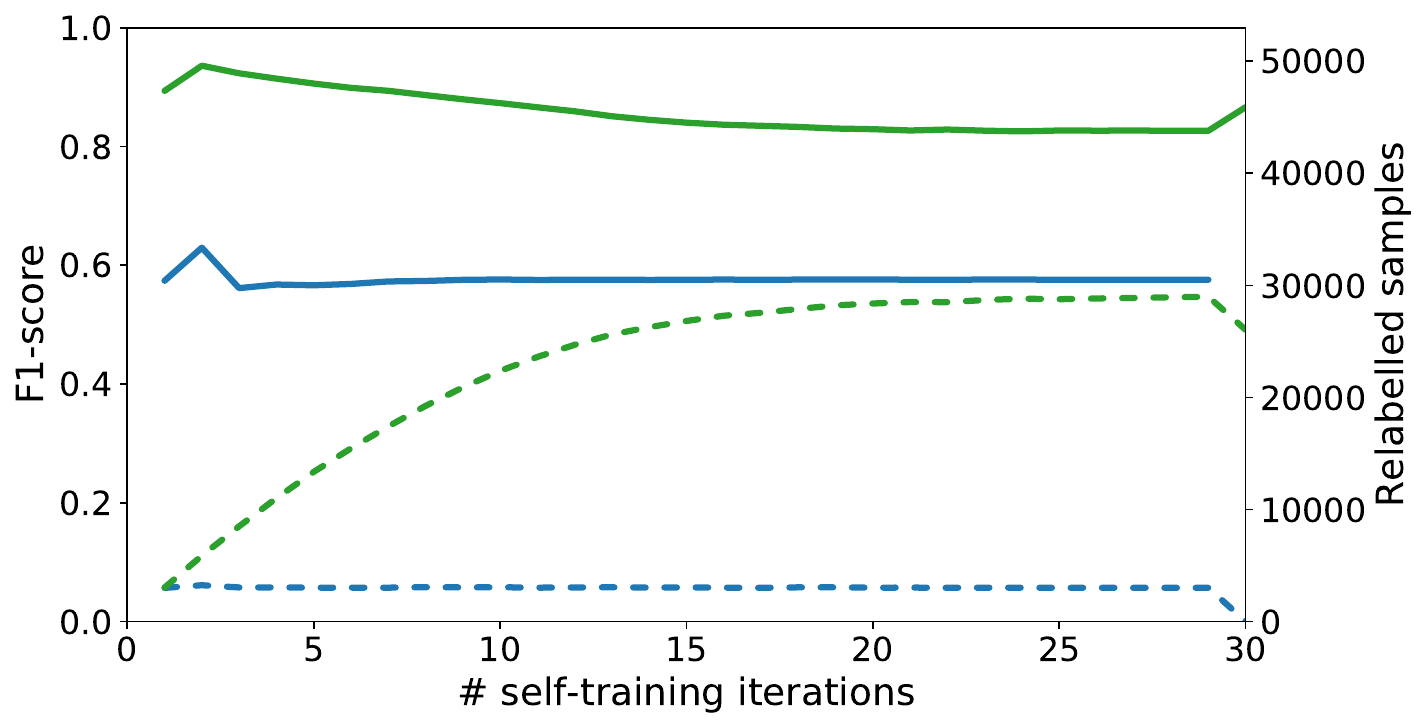}
		\caption{Multi-positive-unlabelled.}  \label{subfig:self_training_multipositive}
	\end{subfigure}
	%
	\caption{SETRED with and without decoupled start.}  \label{fig:self_training}
\end{figure}

In figure \ref{subfig:self_training_semisupervised} we plot the number of relabelled samples (dashed) as well as the F1-score (solid) of the relabelled samples, when using SETRED on the semi-supervised problem. SETRED can clearly solve this kind of problem with high performance. 
In figure \ref{subfig:self_training_pu} we run SETRED on the positive-unlabelled problem, with two different settings. The blue curve is SETRED running out of the box, attempting to predict both positive classes and the negative class. The first iteration selects more samples from the negative class to counter the initial label-imbalance. SETRED has a hard time converging to a consistent model and does not manage to relabel many samples. 
The green curve starts out by optimizing the initial decoupling problem which allows the model to start off better. The rest of the curve is normal SETRED which keeps performance around 0.80 F1-score and manages to label all unlabelled samples accordingly.  
In figure \ref{subfig:self_training_multipositive} we run the same experiment, but on the multi-positive-unlabelled problem. The vanilla version of SETRED has a tough time correctly labelling the unlabelled set and remains in some local minima. Using decoupling to initialize SETRED makes the performance increase greatly and SETRED takes it from there keeping a relatively high performance (it does though only manage to label about half of the samples). \\

\renewcommand{\temp}{0.25}

\section{Conclusion}

We present a principled, unified approach to non-standard classification, which we call \textit{decoupling}. The framework allows inferring class-distributions $p(y \given \vx)$ from label distributions $p(s \given \vx)$, which can be learning by machine learning models. 
We derive the approximations needed for the optimization of the framework and show how it operates on simulated data and on the Fashion MNIST dataset. We show how the framework assists the semi-supervised learning method SETRED in operating on positive-unlabelled and multi-positive-unlabelled learning problems.


\clearpage
\bibliography{ref/misinfo,ref/ML,ref/semi_supervised,ref/pu,ref/noisy_labels,ref/attacks}


\clearpage

\appendix
\part*{Supplementary Material}

\setcounter{equation}{0}
\renewcommand{\theequation}{S.\arabic{equation}}


\section{Positive-Unlabelled Learning}  \label{append:positive_unlabelled_learning_elkan}

\subsection{Elkan and Noto 2008}
In \cite{elkan_learning_2008} they prove the following lemma and proof.
Say we have a positive class and a negative class, and that the probability of incorrectly selecting a negative as a positive is 0. \\
The probability of a selection is
\begin{align*}
	p(s = 1 \given \vx) &= p(s = 1 \given y = 1) p(y = 1 \given \vx) \\
	&+ p(s = 1 \given y = 0) p(y = 0 \given \vx) \\
	&= p(s = 1 \given y = 1) p(y = 1 \given \vx) \\
	&+ 0 \cdot p(y = 0 \given \vx) \\
	&= p(s = 1 \given y = 1) p(y = 1 \given \vx) \\
\end{align*}
So therefore
\begin{align}
	p(y = 1 \given \vx) &= \frac{p(s = 1 \given \vx)}{\rho} \label{eq:elkan_main_lemma} \\
	\rho = p(s &= 1 \given y = 1) \nonumber
\end{align}

\cite{elkan_learning_2008} further concludes that
\begin{align}
	p(s = 1 \given \vx) \leq \rho \label{eq:elkan_p(s)_ceiling}
\end{align} 
in order for the probabilities to remain well behaved after scaling.

\subsection{Positive-Unlabelled Learning with Decoupling}
\renewcommand{\temp}{0.3}
\begin{figure}[b]
	\centering
	\includegraphics[scale=\temp]{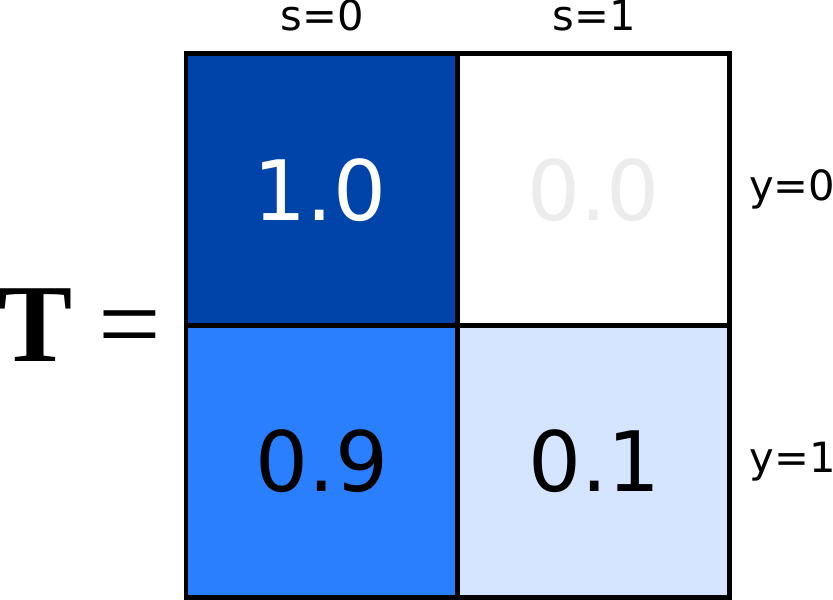}
	\caption{Example of a transition matrix for positive-unlabelled learning.} 
	\label{append_fig:special_case_pu}
\end{figure}

Say we want to solve the decoupling problem for positive unlabelled learning
\begin{align}
	\mS = \mY \mT,
\end{align}
by using the inverse of $\mT$. \\

In the decoupling framework the positive-unlabelled transition matrix $\mT$ is (example in figure \ref{append_fig:special_case_pu})
\begin{align*}
	\mT = \begin{bmatrix}
	1.0 & 0.0 \\
	1 - \rho & \rho 
	\end{bmatrix}  \qquad \qquad \rho = p(s = 1 \given y = 1).
\end{align*}

The inverse for a 2-by-2 matrix is
\[
	\mA = \begin{bmatrix}
	a & b \\ c & d
	\end{bmatrix} \qquad \qquad
	\mA^{-1} = \frac{1}{ad - bc} \begin{bmatrix}
	d & -b \\ -c & a
	\end{bmatrix}.
\]

Therefore the inverse of $\mT$ is
\begin{align*}
	\frac{1}{ad - bc} = \frac{1}{1 \cdot \rho - 0 \cdot (1 - \rho)} = \frac{1}{\rho} 
	\end{align*}\begin{align*}
	\begin{bmatrix}
		d & -b \\ -c & a
	\end{bmatrix} 
	= 
	\begin{bmatrix}
		\rho & 0 \\ \rho - 1 & 1
	\end{bmatrix} 
	\end{align*}\begin{align*}
	\mT^{-1} = 
	\frac{1}{\rho}
	\begin{bmatrix}
		\rho & 0 \\ \rho - 1 & 1
	\end{bmatrix} = 
	\begin{bmatrix}
		1 & 0 \\ \frac{\rho - 1}{\rho} & \frac{1}{\rho}
	\end{bmatrix}
\end{align*}

The distribution across classes for a sample $\vx$ is (transposed for ease of reading)
\begin{align*}
	\mY^{\top} 
		&= \big( \mS \mT^{-1} \big)^{\top} \\
		&= 
		\left( 
		\Big[ \: p( s= 0 \given \vx) \;\;\; p( s= 1 \given \vx) \: \Big]
		\begin{bmatrix}
			1 & 0 \\ \frac{\rho - 1}{\rho} & \frac{1}{\rho}
		\end{bmatrix} 
		\right)^{\top} \\
		&= 
		\begin{bmatrix}
			p( s= 0 \given \vx) + p( s= 1 \given \vx) \cdot \frac{\rho - 1}{\rho} \\
			p( s= 1 \given \vx) \cdot \frac{1}{\rho}
		\end{bmatrix} \\
		&= 
		\begin{bmatrix}
			p(y = 0 \given \vx) \\
			p(y = 1 \given \vx)
		\end{bmatrix}.
\end{align*}
We see that the probability of $y=1$ is $\frac{1}{\rho} \cdot p( s= 1 \given \vx)$ like in Elkan and Noto's result in \ref{eq:elkan_main_lemma}. The constraint of \ref{eq:elkan_p(s)_ceiling} comes naturally from this result, but can also be showed from $p(y = 0 \given \vx)$ together with the corresponding constraint on $p(s=0 \given \vx)$ by \\
\begin{align*}
	0 &\leq p(y = 0 \given \vx) \\
	0 &\leq p( s= 0 \given \vx) + p( s= 1 \given \vx) \cdot \frac{\rho - 1}{\rho} \\
	0 &\leq 1 - p(s = 1 \given \vx) + p(s = 1 \given \vx) \cdot \frac{\rho - 1}{\rho} \\
	0 &\leq p(s = 1 \given \vx) \cdot \left(\frac{\rho - 1}{\rho} - \frac{\rho}{\rho}\right) + 1 \\
	-1 &\leq -\frac{1}{\rho} \: p(s = 1 \given \vx) \\
	1 &\geq \frac{1}{\rho} \: p(s = 1 \given \vx) \\
	\rho &\geq p(s = 1 \given \vx) \\[2mm]
	p(s = 1 &\given y = 1) \geq p(s = 1 \given \vx).
\end{align*}

Or alternatively
\begin{align*}
	0 &\leq p(y = 0 \given \vx) \\
	0 &\leq (1 - p(s = 0 \given \vx)) \cdot \frac{\rho - 1}{\rho} + p(s = 0 \given \vx) \\
	0 &\leq \frac{\rho - 1}{\rho} - \frac{\rho - 1}{\rho} p(s = 0 \given \vx) + p(s = 0 \given \vx) \\
	0 &\leq \frac{\rho - 1}{\rho} + \left(\frac{\rho}{\rho} - \frac{\rho - 1}{\rho}\right) p(s = 0 \given \vx) \\
	0 &\leq \frac{\rho - 1}{\rho} + \frac{1}{\rho} p(s = 0 \given \vx) \\
	1 - \rho &\leq p(s = 0 \given \vx) \\[2mm]
	p(s = 0 &\given y = 1) \leq p(s = 0 \given \vx).
\end{align*}

\section{Evidence Lower Bound (ELBO)}  \label{appendix:elbo_derivation}

One approach to variational inference is to optimize the evidence lower bound - ELBO, given by
\begin{align}
	\text{ELBO}(q) &= \E_q[ \ln p(\vz, \vx) ] - \E_q[ \ln q(\vz) ] \\
		&= 
			\underbrace{\E_q[ \ln p(\vx \given \vz) ]}_{\text{Likelihood term}} 
			+ \underbrace{\E_q[ \ln p(\vz) ]}_{\text{Prior term}} 
			- \underbrace{\E_q[ \ln q(\vz) ]}_{\text{Entropy term}}. \nonumber
\end{align}
where $q$ is the variational distribution and $p$ is the true distribution. \\

For our problem this becomes
\begin{align}
	\text{ELBO} 
		&= \E_q[ \ln p(\mS \given \mY, \mT) ] 
			+ \E_q[ \ln p(\mY \given \vgamma) \: p(\mT) \: p(\vgamma) ] \nonumber \\
		&\qquad - \E_q[ \ln q(\mY) \: q(\mT) \: q(\vgamma) ]. \label{eq:problem_elbo}
\end{align}

\subsection{Likelihood term}
Consider the first expectation of the ELBO (\ref{eq:problem_elbo})
\begin{align}
	\E_q[ \ln p(\mS \given \mY&, \mT) ] 
		= \int_{q\triangle} \ln p(\mS \given \mY, \mT) \d \mT \d \mY \\
		&= \int_{q\triangle} \ln \bigg( \prod_{is} (\mY \mT)_{is}^{\emS_{is}} \bigg) \d \mT \d \mY \nonumber \\
		&= \int_{q\triangle} \sum_{is} \emS_{is} \ln \big( (\mY \mT)_{is} \big) \d \mT \d \mY \nonumber \\
		&= \sum_{is} \emS_{is} \int_{q\triangle} \ln \Big( \sum_{y} \emY_{iy} \emT_{ys} \Big) \d \mT \d \mY, \nonumber 
\end{align}
where the integrals are over the simplex according to the $q$-distributions. We will show how to approximate this integral in section \ref{appendix:expected_log_approximation_for_elbo_likelihood}.

\subsection{Prior term}
The second expectation of the ELBO (\ref{eq:problem_elbo}) is
\begin{align}
	\E_q[ \ln p(\mY \given \vgamma) \: p(&\mT) \: p(\vgamma) ] 
		= \E_q[ \ln p(\mY \given \vgamma) ] \label{eq:problem_elbo_second_term} \\[1mm]
		& + \E_q[ \ln p(\mT) ] 
			+ \E_q[ \ln p(\vgamma) ]. \nonumber
\end{align}

We handle each of these expectations separately, starting with the expectation of $\ln p(\mY \given \vgamma)$
\begin{align}
	\E_q&\big[ \ln p(\mY \given \vgamma) \big] 
		= \E_q\Bigg[ \ln \bigg( \frac{1}{B(\vgamma)} \prod_{iy} \emY_{i y}^{\evgamma_y - 1} \bigg) \Bigg] \nonumber \\
		&= \E_q\bigg[ \sum_{iy} (\evgamma_y - 1) \ln \emY_{i y} \bigg] - \E_q[\ln B(\vgamma)] \\
		&= \sum_{iy} \big(\E_q[\evgamma_y] - 1\big) \E_q\big[ \ln \emY_{i y} \big] - \E_q[\ln B(\vgamma)]. \nonumber
\end{align}

Expectation of $\ln p(\mT)$
\begin{align}
	\E_q[ \ln p(\mT) ]
		&= \E_q\Bigg[ \ln \bigg( \frac{1}{B(\mA)} \prod_{ys} \emT_{ys}^{\emA_{ys} - 1} \bigg) \Bigg] \\
		&= \E_q\bigg[ \sum_{ys} (\emA_{ys} - 1) \ln \emT_{ys} \bigg] - \ln B(\mA) \nonumber \\
		&= \sum_{ys} (\emA_{ys} - 1) \E_q\big[ \ln \emT_{ys}\big] - \ln B(\mA). \nonumber
\end{align}

Expectation of $\ln p(\vgamma)$
\begin{align}
	\E_q[& \ln p(\vgamma) ] 
		= \E_q\Bigg[ \ln \bigg( \frac{1}{Z(\eta, \vlambda) B(\vlambda)^{\eta}} \e^{- \sum_y \evlambda_y \evgamma_y} \bigg) \Bigg] \nonumber \\
		&= - \E_q\bigg[ \sum_y \evlambda_y \evgamma_y \bigg] - \ln \Big( Z(\eta, \vlambda) B(\vlambda)^{\eta} \Big) \nonumber \\
		&= - \sum_y \evlambda_y \E_q\big[ \evgamma_y \big] - \ln \Big( Z(\eta, \vlambda) B(\vlambda)^{\eta} \Big). 
\end{align}

Thus (\ref{eq:problem_elbo_second_term}) becomes
\begin{align}
	\E_q[ &\ln p(\mY \given \vgamma) \: p(\mT) \: p(\vgamma) ] \\
		&= \sum_{iy} \big(\E_q[\evgamma_y] - 1\big) \E_q\big[ \ln \emY_{i y} \big] - \E_q[\ln B(\vgamma)] \nonumber \\
		&\qquad + \sum_{ys} (\emA_{ys} - 1) \E_q\big[ \ln \emT_{ys} \big] - \ln B(\mA) \nonumber \\
		&\qquad - \sum_y \evlambda_y \E_q\big[ \evgamma_y \big] - \ln \Big( Z(\eta, \vlambda) B(\vlambda)^{\eta} \Big) \nonumber \\
		&= \mathlarger{\mathlarger{\sum}}_y \bigg(
			\sum_i \big(\E_q[\evgamma_y] - 1\big) \E_q\big[ \ln \emY_{i y} \big] \nonumber \\
			&\qquad\quad+ \sum_s (\emA_{ys} - 1) \E_q\big[ \ln \emT_{ys} \big] 
			- \evlambda_y \E_q\big[ \evgamma_y \big]
		\bigg) \nonumber \\
		&\qquad - \E_q[\ln B(\vgamma)] - \ln B(\mA) - \ln \Big( Z(\eta, \vlambda) B(\vlambda)^{\eta} \Big). \nonumber
\end{align}

The weighted sum of expected elements from a Dirichlet distribution can be found in (\ref{eq:dirichlet_w_sum_expectation}), with its derivative in (\ref{eq:dirichlet_w_sum_expectation_derivative}). 
The weighted sum of expected log-elements from a Dirichlet distribution can be found in (\ref{eq:dirichlet_w_sum_log_expectation}), with its derivative in (\ref{eq:dirichlet_w_sum_log_expectation_derivative}).

\subsection{Entropy term}
The third expectation of the ELBO (\ref{eq:problem_elbo}) is
\begin{align}
	\E_q[ &\ln q(\mY) \: q(\mT) \: q(\vgamma) ]  \\[1mm]
		&=     \E_q[ \ln q(\mY) ]
			+ \E_q[ \ln q(\mT) ]
			+ \E_q[ \ln q(\vgamma) ] \nonumber \\[1mm]
		&=    \sum_i \E_q[ \ln q(\mY_{i:}) ]
			+ \sum_y \E_q[ \ln q(\mT_{y:}) ]
			+ \E_q[ \ln q(\vgamma) ] \nonumber \\
		&=    \sum_i H_{\dirichlet}(\mY_{i:})
			+ \sum_y H_{\dirichlet}(\mT_{y:})
			+ H(\vgamma), \nonumber
\end{align}
which is a sum of the entropies of each variational distribution. The entropies of the Dirichlet distributions are known analytically (\ref{eq:dirichlet_entropy}), with its derivative in (\ref{eq:dirichlet_entropy_derivative}). 
\vfill 

\subsection{ELBO Components}
Finally the ELBO becomes
\begingroup 
\setlength{\abovedisplayskip}{5pt}
\setlength{\belowdisplayskip}{2pt}
\setlength{\abovedisplayshortskip}{0pt}
\setlength{\belowdisplayshortskip}{0pt}
\begin{align}
	\text{ELBO} 
		&= \E_q[ \ln p(\mS \given \mY, \mT) ] 
			+ \E_q[ \ln p(\mY \given \vgamma) \: p(\mT) \: p(\vgamma) ] \nonumber \\[1mm]
			&- \E_q[ \ln q(\mY) \: q(\mT) \: q(\vgamma) ] \label{append:eq_final_elbo_components}
\end{align}
\begin{align}
	&= \sum_{is} \emS_{is} \int_{q\triangle} \ln \Big( \sum_{y} \emY_{iy} \emT_{ys} \Big) \d \mT \d \mY
	 &&{\color{sRed}\bm{\bigg\}}} \: \text{(L)} \nonumber \\
		&\qquad +
			\sum_{iy} \big(\E_q[\evgamma_y] - 1\big) \E_q\big[ \ln \emY_{i y} \big]
		 &&{\color{s_sek_blue}\bm{\bigg\}}} \: \text{(P)} \nonumber \\
		&\qquad + \sum_{ys} (\emA_{ys} - 1) \E_q\big[ \ln \emT_{ys} \big]
			- \sum_y \evlambda_y \E_q\big[ \evgamma_y \big] 
		 \hspace{-3.5mm}&&{\color{s_sek_blue}\bm{\bigg\}}} \: \text{(P)} \nonumber \\
		&\qquad 
   			- \sum_i H_{\dirichlet}(\mY_{i:})
			- \sum_y H_{\dirichlet}(\mT_{y:})
		 &&{\color{s_sek_blue}\bm{\bigg\}}} \: \text{(E)} \nonumber \\
		&\qquad
			- H_{q}(\vgamma) - \E_q[\ln B(\vgamma)] 
		 &&{\color{sRed}\bm{\bigg\}}} \: \text{(Cl)} \nonumber \\
		&\qquad - \ln B(\mA) - \ln \Big( Z(\eta, \vlambda) B(\vlambda)^{\eta} \Big),
	 &&{\color{s_sek_green}\bm{\bigg\}}} \: \text{(Co)} \nonumber
\end{align}
\endgroup
\begin{description}
	\item[(L)] Likelihood which is problematic
	\item[(P)] Priors with analytic solutions
	\item[(E)] Entropies with analytic solutions
	\item[(Cl)] Class prior, unknown
	\item[(Co)] Constants
\end{description}
where the constants can be omitted for optimization purposes.

\subsection[Variational Distribution of gamma]{Variational Distribution of $\vgamma$}

The ELBO in (\ref{append:eq_final_elbo_components}) illustrated the required properties for the variational distribution needed for $\vgamma$. It needs to be in the domain of $[0, \infty]$ and we must be able to compute the following
\begin{itemize}
	\item The expectation of one element: $\E_q[\evgamma_y]$
	\item The Entropy of the distribution: $\E_q[\ln p(\vgamma)]$
	\item The expectation logarithm of the multivariate binomial coefficient: $\E_q[ \ln B(\vgamma) ]$
\end{itemize}
While the domain and the first two quantities are quite common and should be easy to find, the last one is more troublesome. \\

We leave it for future work to find a suitable variational distribution for $\vgamma$.
We do note though that if we find a distribution which satisfy the first requirements (several distributions are available), then we could sample $\E_q[ \ln B(\vgamma) ]$, which is a fairly cheap operation.

\subsection[ELBO Without Distribution on gamma]{ELBO Without Distribution on $\vgamma$}  \label{appendix:elbo_with_constant_gamma}

If we instead let $\vgamma$ (corresponding to known class prior with known certainty) be a constant, the ELBO becomes
\begingroup 
\setlength{\abovedisplayskip}{5pt}
\setlength{\belowdisplayskip}{2pt}
\setlength{\abovedisplayshortskip}{0pt}
\setlength{\belowdisplayshortskip}{0pt}
\begin{align}
	\text{ELBO} 
		&= \E_q[ \ln p(\mS \given \mY, \mT) ] 
			+ \E_q[ \ln p(\mY \given \vgamma) \: p(\mT) \: p(\vgamma) ] \nonumber \\[1mm]
			&- \E_q[ \ln q(\mY) \: q(\mT) \: q(\vgamma) ] \label{append:eq_final_elbo} 
\end{align}
\begin{align}
	&= \sum_{is} \emS_{is} \int_{q\triangle} \ln \Big( \sum_{y} \emY_{iy} \emT_{ys} \Big) \d \mT \d \mY
	 &&{\color{sRed}\bm{\bigg\}}} \: \text{(L)} \nonumber \\
		&\qquad +
			\sum_{iy} (\evgamma_y - 1) \E_q\big[ \ln \emY_{i y} \big]
		 &&{\color{s_sek_blue}\bm{\bigg\}}} \: \text{(P)} \nonumber \\
		&\qquad + \sum_{ys} (\emA_{ys} - 1) \E_q\big[ \ln \emT_{ys} \big]
		 \hspace{-3.5mm}&&{\color{s_sek_blue}\bm{\bigg\}}} \: \text{(P)} \nonumber \\
		&\qquad 
   			- \sum_i H_{\dirichlet}(\mY_{i:})
			- \sum_y H_{\dirichlet}(\mT_{y:})
		 	&&{\color{s_sek_blue}\bm{\bigg\}}} \: \text{(E)} \nonumber \\
		&\qquad
			- \ln B(\vgamma) - \sum_y \evlambda_y \evgamma_y
		 	&&{\color{s_sek_green}\bm{\bigg\}}} \: \text{(Co)} \nonumber \\
		&\qquad - \ln B(\mA) - \ln \Big( Z(\eta, \vlambda) B(\vlambda)^{\eta} \Big).
	 		&&{\color{s_sek_green}\bm{\bigg\}}} \: \text{(Co)} \nonumber
\end{align}
\endgroup

\section{Expected Logarithm Approximation}  \label{appendix:expected_log_approximation}

The Taylor series of a function $f(x)$ is around a point $x=a$
\begin{align}
	f(x) &\approx f(a) + \sum_{p=1}^K \frac{f^{(p)}(a)}{p!} (x - a)^p
\end{align}

The derivative of the logarithm is
\begin{align}
	\diff[^p]{x^p} \ln(x) = (-1)^{p - 1} \frac{1}{x^p} (p - 1)!
\end{align}

The Taylor series of the natural logarithm therefore becomes
\begin{align}
	&\qquad\qquad\text{around~}~ x=a \\
	\ln(x) &\approx
		\ln(a) + \sum_{p=1}^K
			\underbrace{ (-1)^{p - 1} \frac{1}{a^p} (p - 1)! }_{ \diff[^p]{x^p} \ln(x) \big|_{x=a} }
			\frac{1}{p!} (x - a)^p \nonumber \\
		&= \ln(a) + \sum_{p=1}^K (-1)^{p - 1} \frac{1}{p \cdot a^p} (x - a)^p \nonumber
\end{align}

We now assume that $x$ is a stochastic variable ($X$) and set the approximation point to be equal to $X$'s expectation: $a = E[X] = \mu$

\begin{align}
	&\qquad\qquad \text{around~}~ x=\mu \\
	\ln(X) 
		&\approx \ln(E[X]) + \sum_{p=1}^K (-1)^{p - 1} \frac{1}{p \cdot E[X]^p} (X - E[X])^p \nonumber \\
		&= \ln(\mu) + \sum_{p=1}^K (-1)^{p - 1} \frac{1}{p \cdot \mu^p} (X - \mu)^p. \nonumber
\end{align}

The expectation of this approximation becomes
\begin{align}
	\E[\ln(X)]
		&= \ln(\mu) + \sum_{p=1}^K (-1)^{p - 1} \frac{1}{p \cdot \mu^p} \E\Big[ (X - \mu)^p \Big] \nonumber \\
		&= \ln(\mu) + \sum_{p=1}^K (-1)^{p - 1} \frac{1}{p \cdot \mu^p} \: \mu_p,
\end{align}
where $\mu_p$ is the $p$'th moment of $X$ (section \ref{append:moments}).

\section{Expansions of Sums}  \label{appendix:expansions_of_sums}
For the following approximation we use the notion of the following set
\begin{align}
	\mathcal{B}^{(n,k)} = \Big\{\: \vbeta \in \big(\N \cup \{0\}\big)^{n} \::\: \vone^{\top}\vbeta = k \:\Big\}, 
\end{align}
which for given natural numbers $k \in \N$ and $n \in \N$ is the set of non-negative, integer-valued vectors of length $n$, which sums to $k$. 
It is useful for specifying all polynomial terms of order $k$, with variables from a vector $\vx$ of length $n$, as
\begin{align}
	\sum_{ \vbeta \in \mathcal{B}^{(n,k)} } \!\! \prod_i^{n} \evx_i^{\vbeta_i}.
\end{align}
This corresponds to the sum of the $k$-th Cartesian product of the vector onto itself.

We will furthermore use the following rearranging of sums ($i_0 \leq j_0 \leq n$)
\begin{align}
	\sum_{i=i_0}^{n} \sum_{j=j_0}^{i} \emX_{ij} 
	&= \sum_{j=j_0}^{n} \sum_{\substack{i=\\\max(i_0, j)}}^{n} \emX_{ij}.
	\label{eq:inner_out_sum_rearrange}
\end{align}

\subsection{Power of Sum}  \label{appendix:power_of_sum}

\begin{figure}
	\centering
	\begin{subfigure}{0.9\linewidth}
		\centering
		\includegraphics[width=\linewidth, page=1]{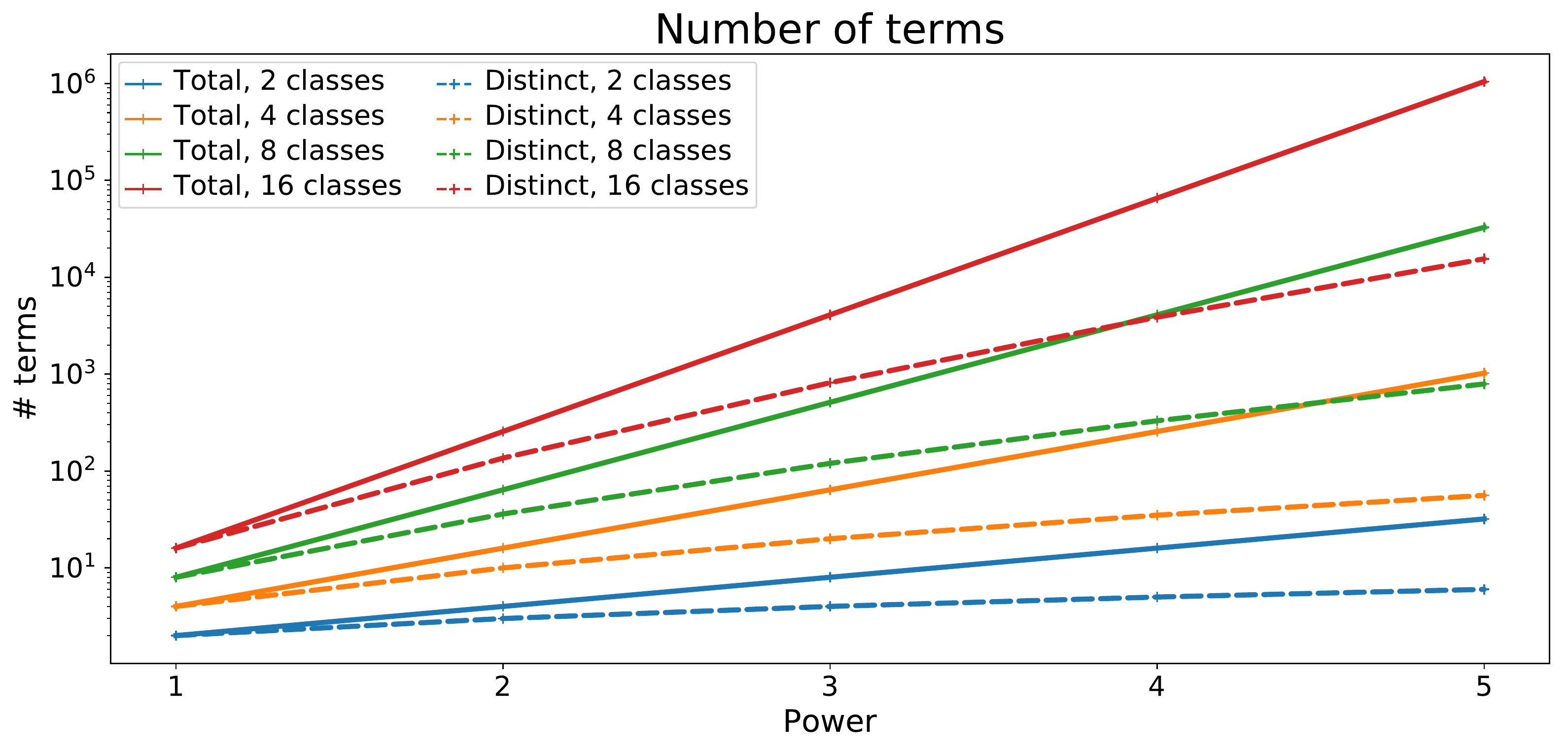}
		\caption{Number of term and number of distinct terms from power-of-sum. Note the log-scale.}
	\end{subfigure} \\[3mm]
	\begin{subfigure}{0.9\linewidth}
		\centering
		\includegraphics[width=\linewidth, page=2]{fig/python/term_complexities.pdf}
		\caption{Ratio between number of terms and distinct terms from power-of-sum.}
	\end{subfigure}
	\caption{Comparison of complexities. } \label{append_fig:complexities}
\end{figure}

Consider the $p$'th power of a sum
\begin{align}
	\bigg(\sum_k^K \evx_k\bigg)^p.
\end{align}
If we expand this parenthesis it will produce $K^p$ terms, but due to the commutative property of multiplication many of these terms will be identical. All of the terms will be composed by exactly $p$ factors. The number of distinct terms is equal to the number of combinations of the $\vx$-elements, with replacement
\begin{align}
	N_c 
		= \frac{\Gamma(K + p)}{\Gamma(p + 1)\Gamma(K)}.
\end{align}
This term is smaller than $K^p$ as illustrated in figure \ref{append_fig:complexities}. \\

We can write any of these terms as
\begin{align}
	\prod_i \evx_i^{\vbeta_i}, \qquad \vbeta \in \mathcal{B}^{(K, p)}.
\end{align}

The number of terms described by a exponent-vector $\vbeta$ is the number of ways we can select those exponents from $p$ products. We can thus write the power of the sum as
\begin{align}
	\bigg(\sum_k^K \evx_k\bigg)^p 
	= \!\!\!\sum_{ \vbeta \in \mathcal{B}^{(K, p)} } \!\!\! {p \choose \vbeta} \prod_i \evx_i^{\vbeta_i}
	= \xi(\vx, p).
	\label{eq_appendix:power_of_sum}
\end{align}
We will use $\xi(\vx, p)$ to express the sum of all products of order $p$ of elements from $\vx$. Note that
\begin{align}
	\xi(\vx, 0) = 1
\end{align}
as there is one $\vbeta$-vector with sum zero, which will set all elements to the power of zero and product a product of ones.

\subsection{Series Expansion of Function of Sum}  \label{appendix:series_expansion_of_function_of_sum}

Consider a situation where we have a function $f(x)$, which is either intractable, not integrable, not differentiable or otherwise somehow problematic. Assume this function is applied to a sum $\sum_k^K \evx_k$ of element which we also cannot directly compute; for example because we wish to integrate over the function with respect to the variables of the sum. Finally assume that we have the following series approximation
\begin{align}
	f(x) \approx \sum_{p=1}^P \evc_p x^p,
\end{align}
with known coefficients $\evc_p$. \\

We approximate the function of the sum by
\begin{align}
	f\bigg( \sum_k^K \evx_k \bigg) \approx \sum_{p=1}^P \evc_p \bigg(\sum_k^K \evx_k\bigg)^p.
\end{align}

By using (\ref{eq_appendix:power_of_sum}) we get
\begin{align}
	f\bigg( \sum_k^K \evx_k \bigg) &\approx \sum_{p=1}^P \evc_p \xi(\vx, p).
		\label{eq:series_approx_function_betas}
\end{align}

Consider the case where $\evx_0$ is for some reason special. We split the sum over beta-vectors to bring out this element (using (\ref{eq:inner_out_sum_rearrange}))
\begin{align}
	f\bigg( \sum_{k=1}^K \evx_k + \evx_0 \bigg) 
		&\approx \sum_{p=1}^P \evc_p \xi\big(\vx \cup \{\evx_0\}, p\big)
		\label{eq:series_approx_function_betas_x0} \\
		&= \sum_{p=1}^P \evc_p
			\sum_{\ell=0}^p \evx_0^{p - \ell} {p \choose \ell}
			\xi(\vx, \ell) \nonumber \\
		&= \sum_{\ell=0}^P \sum_{\mathclap{\substack{p=\\\max(\ell,1)}}}^P 
			\evc_p \evx_0^{p - \ell} {p \choose \ell} \xi(\vx, \ell), \nonumber
\end{align}
where $\vx$ is the vector without element $\evx_0$, and $\evc_p \evx_0^{p - \ell} {p \choose \ell}$ can be considered a new coefficient.

\section{Expected Logarithm for ELBO Likelihood Term}  \label{appendix:expected_log_approximation_for_elbo_likelihood}
\subsection{Uncentered Moments}

We wish to estimate 
\begin{align}
	\int_{q\triangle}& \ln \Big( \sum_{y} \emY_{iy} \emT_{ys} \Big) \d \mY \d \mT 
		= \E \bigg[ \ln \Big( \sum_{y} \emY_{iy} \emT_{ys} \Big) \bigg]. \nonumber
\end{align}

For simplifying expressions, we first define the following tensor
\begin{align}
	\tS \in \R^{n \times m_s \times m_y}, \qquad
	\etS_{isy} = \emY_{iy} \emT_{ys},
\end{align}
so that
\begin{align}
	\sum_{y} \emY_{iy} \emT_{ys} = \vone^{\top} \tS_{is:}
\end{align}

We can expand the power of the sum of $\emY$ and $\emT$ to
\begin{align}
	\bigg( \sum_{y} \emY_{iy}& \emT_{ys} \bigg)^p 
		= \bigg( \sum_{y} \etS_{isy} \bigg)^p \\
		%
		&= \!\!\!\sum_{ \vbeta \in \mathcal{B}^{(K, n)} } \!\!\! 
			{n \choose \vbeta} \prod_y \etS_{isy}^{\vbeta_y} 
		= \xi\big(\tS_{is:}, p\big). \nonumber
\end{align}

We can compute the expectation of this expression by
\begin{align}
	\E\Bigg[ \bigg(& \sum_{y} \emY_{iy} \emT_{ys} \bigg)^p \Bigg]
		= \E\Big[ \xi\big(\tS_{is:}, p\big) \Big] \\
		&= \!\!\!\sum_{ \vbeta \in \mathcal{B}^{(K, n)} } \!\!\! 
			{n \choose \vbeta} \E_q\bigg[ \prod_y \etS_{isy}^{\vbeta_y} \bigg] \nonumber \\
		&= \!\!\!\sum_{ \vbeta \in \mathcal{B}^{(K, n)} } \!\!\! 
			{n \choose \vbeta} 
			\E_q\bigg[ \prod_y \emY_{iy}^{\vbeta_y} \bigg]
			\E_q\bigg[ \prod_y \emT_{ys}^{\vbeta_y} \bigg]. \nonumber
\end{align}

We can now express the expectation and the uncetered moments as
\begin{align}
	\mu = \E[X] 
		&= \E\bigg[ \sum_{y} \emY_{iy} \emT_{ys} \bigg]
		= \sum_{y} \E\Big[ \emY_{iy} \emT_{ys} \Big] \\
	\E[X^p] 
		&= \E\Bigg[ \bigg( \sum_{y} \emY_{iy} \emT_{ys} \bigg)^p \Bigg]
		= \E\Big[ \xi\big(\tS_{is:}, p\big) \Big]. \nonumber
\end{align}

\subsection{Central Moments}

The $k$'th central moment is (section \ref{append:moments}, \textit{\nameref{append:moments}})
\begin{align}
	&\mu_k = \E\big[ (X - \mu)^k \big] \\
		&= \sum_{j=0}^k {k \choose j} (-1)^{k - j} \E[X^j] \mu^{k - j} \nonumber \\
		&= \sum_{j=0}^k {k \choose j} (-1)^{k - j} 
			\underbrace{\E\Big[ \xi\big(\tS_{is:}, j\big) \Big]}_{\E[X^j]}
			\bigg( \underbrace{\sum_{y} \E\Big[ \emY_{iy} \emT_{ys} \Big]}_{\mu=\E[X]} \bigg)^{k - j} \nonumber
\end{align}

\subsection{Final Approximation}

By inserting the central moments into the approximation of the natural logarithm (section \ref{appendix:expected_log_approximation}) we find
\begin{align}
	&\E \Big[ \ln \Big( \sum_{y} \emY_{iy} \emT_{ys} \Big) \Big] \\
		&= \ln\Big( \sum_{y} \E\big[ \emY_{iy} \emT_{ys} \big] \Big) 
			+ \sum_{p=1}^K 
				\frac{(-1)^{p - 1}}{p \cdot \Big( \sum_{y} \E\big[ \emY_{iy} \emT_{ys} \big] \Big) ^p} \nonumber \\
				&\:\:\:\: \sum_{j=0}^p {p \choose j} (-1)^{p - j} 
					\E\Big[ \xi\big(\tS_{is:}, j\big) \Big]
					\bigg( \sum_{y} \E\Big[ \emY_{iy} \emT_{ys} \Big] \bigg)^{p - j}
				\nonumber \\
		&= \ln\Big( \sum_{y} \E\big[ \emY_{iy} \emT_{ys} \big] \Big)
			+ \sum_{p=1}^K 
				\sum_{j=0}^p \frac{1}{p} {p \choose j} (-1)^{j - 1} \nonumber \\
					&\qquad\quad \E\Big[ \xi\big(\tS_{is:}, j\big) \Big]
					\bigg( \sum_{y} \E\Big[ \emY_{iy} \emT_{ys} \Big] \bigg)^{p - j - p}
				\nonumber \\	
		&= 
			\ln\Big( \sum_{y} \E\big[ \emY_{iy} \emT_{ys} \big] \Big)
			+ \sum_{p=1}^K  \frac{1}{p} {p \choose 0} (-1)^{0 - 1} \nonumber \\
				&\qquad\quad \E\Big[ \xi\big(\tS_{is:}, 0\big) \Big]
				\bigg( \sum_{y} \E\Big[ \emY_{iy} \emT_{ys} \Big] \bigg)^{-0} \nonumber \\
			&\:\:\:\:+ \sum_{j=1}^K
				\sum_{p=j}^K  \frac{1}{p} {p \choose j} (-1)^{j - 1} \nonumber \\
					&\qquad\quad \E\Big[ \xi\big(\tS_{is:}, j\big) \Big]
					\bigg( \sum_{y} \E\Big[ \emY_{iy} \emT_{ys} \Big] \bigg)^{-j}
				\nonumber \\
		&= 
			\ln\Big( \sum_{y} \E\big[ \emY_{iy} \emT_{ys} \big] \Big)
			- \sum_{p=1}^K  \frac{1}{p} \nonumber \\
			&\qquad+ \sum_{j=1}^K (-1)^{j - 1} 
				\sum_{p=j}^K  \frac{1}{p} {p \choose j} \nonumber \\
					&\qquad\quad
					\bigg( \sum_{y} \E\Big[ \emY_{iy} \emT_{ys} \Big] \bigg)^{-j}
					\E\Big[ \xi\big(\tS_{is:}, j\big) \Big]
				\nonumber \\
		&= \ln\Big( \sum_{y} \E\big[ \emY_{iy} \emT_{ys} \big] \Big)
			- \sum_{p=1}^K  \frac{1}{p} \nonumber \\
			&\qquad+ \sum_{j=1}^K \tau_j 
				\bigg( \sum_{y} \E\Big[ \emY_{iy} \emT_{ys} \Big] \bigg)^{-j} 
				\E\Big[ \xi\big(\tS_{is:}, j\big) \Big] \nonumber
\end{align} \vspace{-3mm}
\begin{align}
	\tau_j &= (-1)^{j - 1} \sum_{p=1}^K  \frac{1}{p} {p \choose j}. \nonumber
\end{align}

%
%

\section{Label-Conditional Class-Probabilities}  \label{appendix:label_conditional_class_probabilities}
Say we get some label-probabilities $\vs$ from an expert. These labels may not uniquely identify the true class, but will probably be very correlated with the correct class. Let's use that to compute the unnormalized posterior
\begin{align}
	p(&y \given s_i, \mX_{i:})
		\appropto \iint_{q\triangle} 
			p(s_i \given y, \mT) \cdot p(y \given \mX_{i:})
			\d\mY \d\mT \nonumber \\[2mm]
		&= \iint_{q\triangle} 
			\emT_{ys_i} \emY_{iy}
			\d\mY \d\mT
		= \E_q[\emT_{ys_i}] \E_q[\emY_{iy}].
\end{align}

By normalizing we get a simple approximation for the label-conditional class probabilities
\begin{align}
	p(y \given s_i, \mX_{i:}) \approx
		\frac{
			\E_q[\emT_{ys_i}] \E_q[\emY_{iy}]
		}{
			\sum_{y'} \E_q[\emT_{y's_i}] \E_q[\emY_{iy'}]
		}.
\end{align}

We arrange the label-conditional class-probabilities in a matrix as follows
\begin{align}
	\mW \in [0, 1]^{n \times m_y}, 
		\quad 0 \leq \mW_{i y} = p(y \given s_i, \vx_i), \\[2mm]
	\mW \vone = \vone, \quad
	\mW_{i y} \approx \frac{
			\E_q[\emT_{ys_i}] \E_q[\emY_{iy}]
		}{
			\sum_{y'} \E_q[\emT_{y's_i}] \E_q[\emY_{iy'}]
		}. \nonumber
\end{align}

\section{Analysis of Agreement Terms}  \label{appendix:agreement_terms}

\subsection[Interpretation of beta-Vectors]{Interpretation of $\vbeta$-Vectors}

Consider the set (as before)
\begin{align}
	\mathcal{B}^{(n,k)} &= \Big\{\: \vbeta \in \big(\N \cup \{0\}\big)^{n} \::\: \vone^{\top}\vbeta = k \:\Big\}.
\end{align}
This set represents all vectors of length $n \in \N$ which sums to $k \in \N$ and only contains natural numbers (and zero). \\
We use it to represent all possible products of a certain order, but we can also interpret it as samples. A beta vector like
\begin{align}
	\vbeta_{\text{ex}} = \begin{bmatrix}
		0 \\ 1 \\ 2 \\ 0
	\end{bmatrix},
\end{align}
represents a sample of size $3$, where one of the samples is of class $2$ and the two of the samples are of class $3$, out of the total $4$ classes. \\

\begin{figure}[h!]
	\centering
	\includegraphics[width=0.6\linewidth]{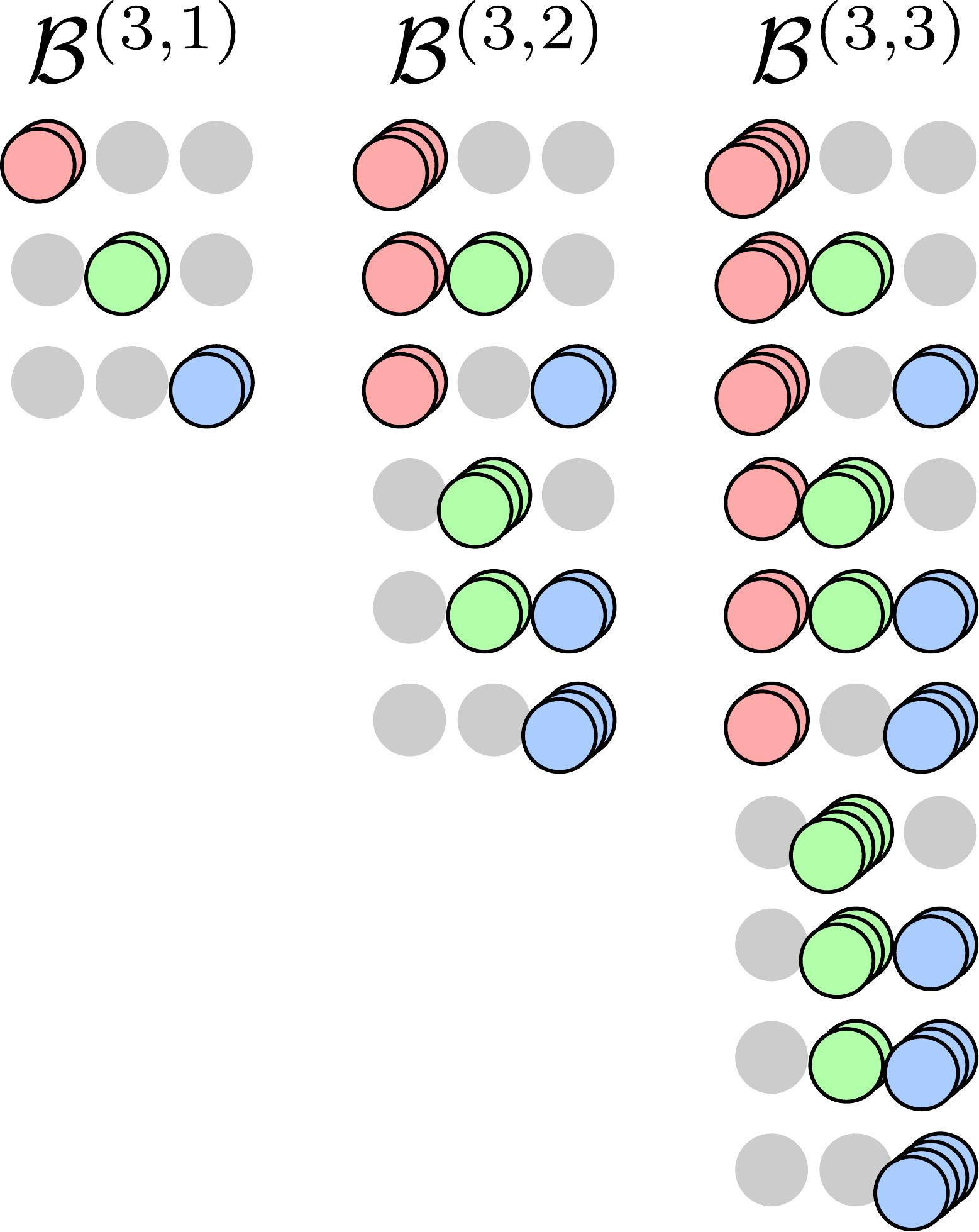}
	\caption{Three examples of $\mathcal{B}^{(n,k)}$ sets. } 
	\label{fig:beta_sets_examples}
\end{figure}

Figure \ref{fig:beta_sets_examples} illustrates three examples of $\mathcal{B}$-sets, all with 3 classes but with 1, 2 and 3 samples, where a sample's class is represented by a coin.

When computing the sum over all $\vbeta$-vectors from a $\mathcal{B}$-set, we can consider it a sum over all possible samples of a specific samples size. The series approximates the expectation by considering all possible samples of a certain size. This makes a approximation methodology reside somewhere between analytical approximations and sampling methods.

\subsection{Agreement}

Let's revisit the agreements of samples and selections; the expectations of $\xi(\tS_{is:}, n)$. We rewrite them with class-probabilities and transition probabilities
\begin{align}
	\E_q&\Big[ \xi\big(\tS_{is:}, n\big) \Big] \\
		&=  \!\!\!\sum_{ \vbeta \in \mathcal{B}^{(K, n)} } \!\!\! 
		{n \choose \vbeta} 
		\E_q\bigg[ \prod_y \emY_{iy}^{\vbeta_y} \bigg]
		\E_q\bigg[ \prod_y \emT_{ys}^{\vbeta_y} \bigg] \nonumber \\
		&= \!\!\!\sum_{ \vbeta \in \mathcal{B}^{(K, n)} } \!\!\! 
		{n \choose \vbeta} 
		\E_q\big[ p( \vy_i = \vbeta \given \mX = \vone \vx_i ) \big] \nonumber \\
		&\qquad\qquad\qquad\:\E_q\big[ p( \vs = s \times \vone \given \vy_i = \vbeta ) \big]. \nonumber
\end{align}
We use $p( \vy_i = \vbeta \given \mX = \vone \vx_i )$ to denote the probability of a dataset of size $n$, all having the same feature vector $\vx_i$ (we use the outer product) and having their classes as denoted in $\vbeta$. $p( \vs = s \cdot \vone \given \vy_i = \vbeta )$ are the transition probabilities from the classes in $\vbeta$ to the selection $s$. \\
We compute the expected probabilities across our variational distributions $q(\cdot)$ and sum them up for all $K^n$ possible terms. \\

The expectation of $\xi(\tS_{is:}, n)$ thus represents the expected agreement between the class-distribution of $\vx_i$ and the selection $s$, for a dataset of $n$ repeated instances of $\vx_i$. Intuitively this seems like a reasonable measure of how well the class-distribution of $\vx_i$ \textit{agrees} with the selection $s$, which leads to their name. \\

\subsection{Example of Agreement}
For investigating how the agreements operate we provide an example.
Table \ref{tbl:agreement_example_Theta} contains the parameters of the selection distributions for two hypothetical classes. The distributions are Dirichlets, so their parameters are positive values, whose sums signifies how peaked the distribution is around its mean (for parameters larger than 1) - how sure we are of the transitions probabilities. One class mostly transitions to selection 1, while the other class transitions to the two selections with equal probability. 
\\

\begin{table}[h!]
	\centering
	\caption{Parameters for example.}
	\begin{subtable}{0.7\linewidth}
		\centering
		\caption{$\dot{\mTheta}$: parameters for the selections distributions of the classes.}
		\label{tbl:agreement_example_Theta}
		\begin{tabular}{lrr}
\toprule
{} &  Selection 1 &  Selection 2 \\
\midrule
Class 1 &           10 &            1 \\
Class 2 &            4 &            4 \\
\bottomrule
\end{tabular}

	\end{subtable} \\[3mm]
	\begin{subtable}{0.7\linewidth}
		\centering
		\caption{$\dot{\mPi}$: parameters for the class distributions of the samples.}
		\label{tbl:agreement_example_Pi}
		\begin{tabular}{lrr}
\toprule
{} &  Class 1 &  Class 2 \\
\midrule
Sample 1       &    10.00 &     1.00 \\
Sample 2       &     1.00 &    10.00 \\
Uniform        &     1.00 &     1.00 \\
Centred        &     4.00 &     4.00 \\
Centred Narrow &    25.00 &    25.00 \\
Multimodal     &     0.35 &     0.35 \\
\bottomrule
\end{tabular}

	\end{subtable}
\end{table}

Table \ref{tbl:agreement_example_Pi} contains the parameters for the class distributions of some hypothetical samples. There are six samples which have all been named as follows
\begin{description}[itemsep=0pt, topsep=4pt]
	\item[Sample 1] A sample that agrees with high probability of class 1.
	\item[Sample 2] A sample that agrees with high probability of class 2.
	\item[Uniform] A sample with uniform distribution. 
	\item[Centred] A sample which is believed to be between the two classes with some certainty.
	\item[Centred Narrow] A sample which is believed to be between the two classes with a lot of certainty (a narrow distribution).
	\item[Multimodel] A sample which is believed to be one of the classes with equal probability, but is\textit{not} believed to be between the classes.
\end{description}

\begin{figure}[h!]
	\centering
	\includegraphics[width=0.8\linewidth, page=4]{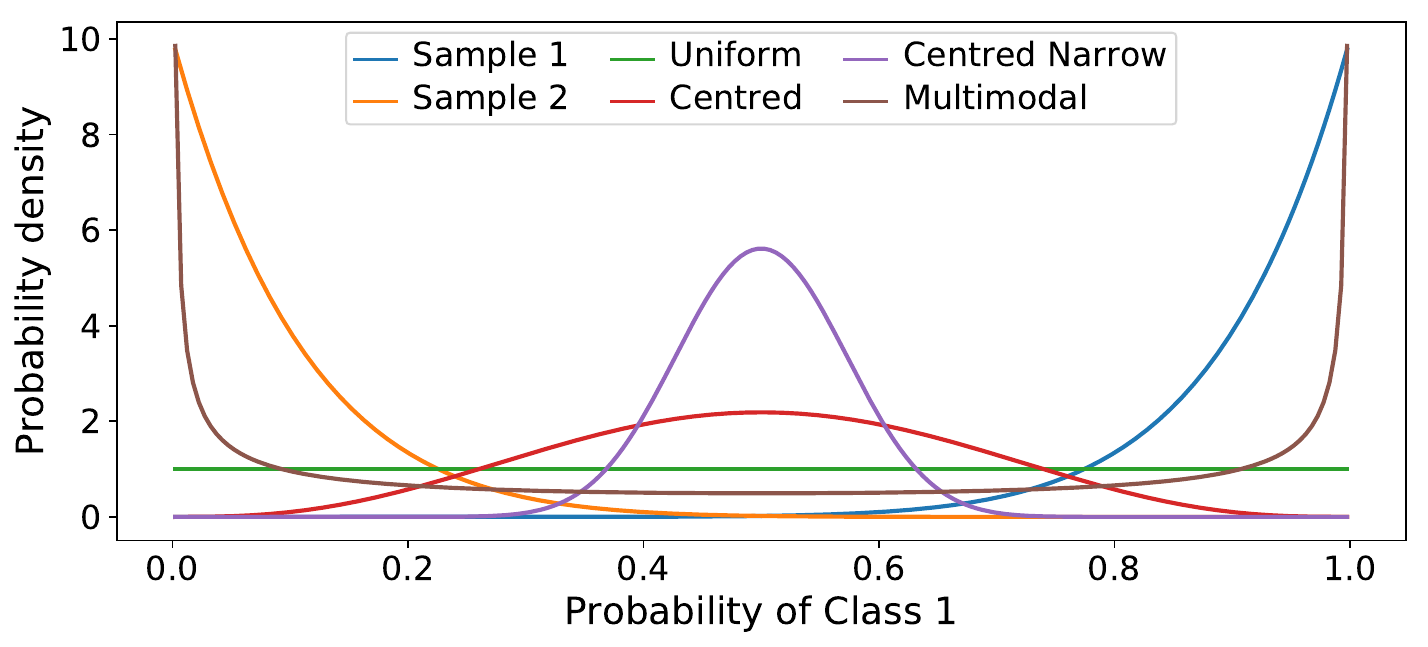}
	\caption{Probability density of the probability of class 1.}
	\label{fig:agreement_example_class_densities}
\end{figure}

We first look at the class-distribution for each sample. For each sample there is some underlying probability mass for each class. Due to uncertainty we don't know these probability masses. Instead we here have a Dirichlet distribution for each sample indicating the probability \textit{density} of each probability \textit{mass} distribution. In Figure \ref{fig:agreement_example_class_densities} we plot the probability density of the probability fo class 1, for each sample.

We see that sample 1 has most of its probability mass close to class 1 and sample 2 has most of its probability mass close to class 2 - as expected. The uniform sample has equal probability density for all distributions between the classes. The centred sample has most of its mass in the center, indicating that this is likely a sample with probability for both classes (intuitively in between the classes). The centred narrow distribution has the same characteristic, but with more certainty for being in between the classes. The multimodal distribution has most of its probability mass close to either of the two classes. This indicates that we believe it has only one class, but we do not know which.  \\

\begin{figure}[h!]
	\centering
	\includegraphics[width=\linewidth, page=2]{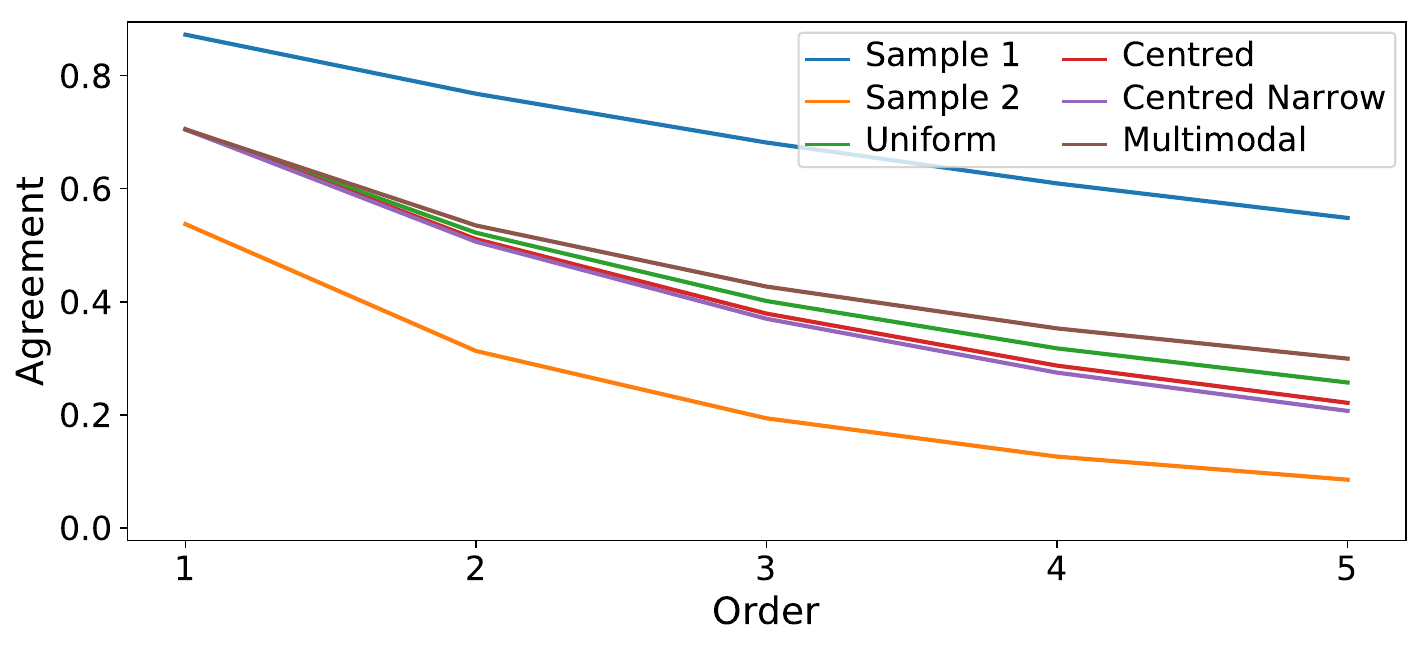}
	\caption{Agreement between each named sample and class 1, for varying order.}
	\label{fig:agreement_example_class_1_agreement}
\end{figure}

Let's now look at the \textit{agreement} term between each sample and class 1. These terms have been plotted in Figure \ref{fig:agreement_example_class_1_agreement} for varying orders.
We first note that in general there is high agreement between sample 1 and class 1, and low agreement between sample 2 and class 1, which is again expected. Of the other four samples we see that the multimodal sample has the second highest agreement. This samples agree well with any class, because it is likely to be any of them. The uniform sample is next in line. This sample also agrees well with all classes, but also has some probability of being shared between classes. The centred and centred narrow samples have lower agreement again, because they are expected to not be dedicated to any class, but rather be a mix. \\

The next important thing to note is that the agreement decreases with order. This makes mathematical sense, as larger sample sizes will have lower probability of all agreeing with one class.  \\

The most important thing is though, that the curves of all four samples with unknown class belongingness have the same agreement for order 1, but are different for all other orders. This is crucial, because it shows why we cannot use a first order approximation - basically we cannot use Jensen's inequality directly on the expectation of the natural logarithm. The first order term cannot capture the differences in certainty of the distributions. \\

We also notice that higher orders separate the agreements more, making them better for determining class belongingness. 

\section{Supervised and Semi-Supervised Learning} \label{append:super_and_semisuper_learning}

\subsection{Supervised Learning}

Let us consider how the decoupling problem handles supervised learning. In supervised learning we have a one-to-one relationship between labels and classes. Each class will always be labelled with their own label and never anything else. The transition matrix for such a problem is a diagonal matrix with ones in the diagonal and zeroes everywhere else. The distribution of the transition matrix will also be a diagonal matrix, where each diagonal element is infinity, corresponding to absolute certainty of class-label transitions (see figure \ref{append_fig:special_case_super}). 

\renewcommand{\temp}{0.25}
\begin{figure}[h!]
	\centering
	\begin{subfigure}{0.45\linewidth}
		\centering
		\includegraphics[scale=\temp]{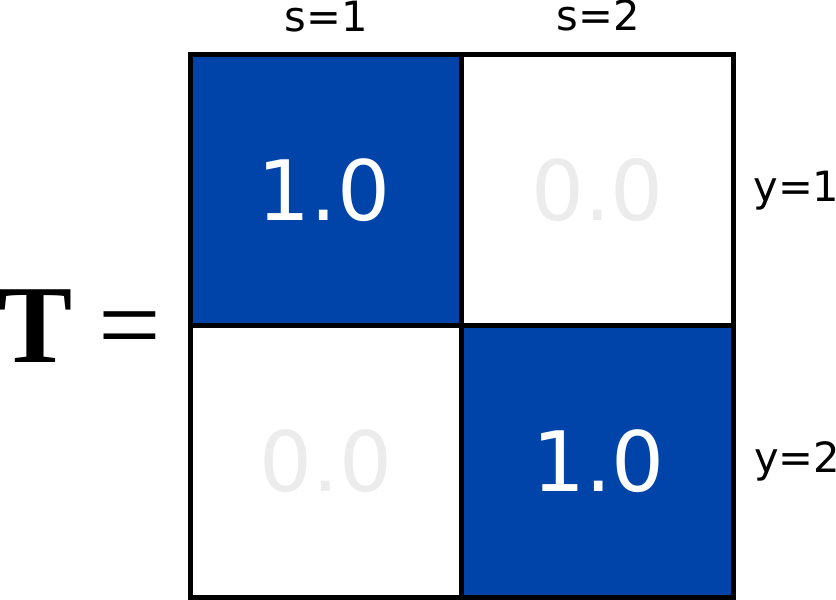}
		\caption{}
	\end{subfigure}
	\begin{subfigure}{0.45\linewidth}
		\centering
		\includegraphics[scale=\temp]{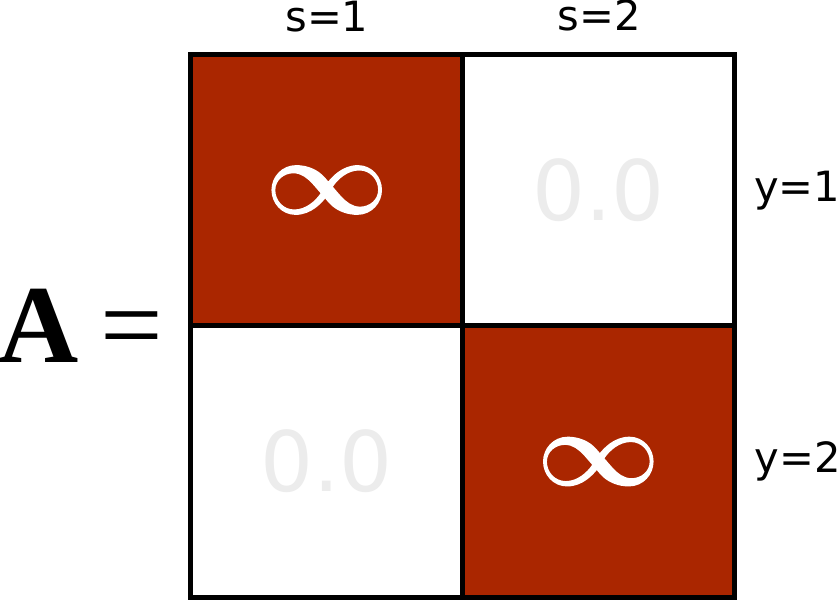}
		\caption{}
	\end{subfigure} \\
	\begin{subfigure}{0.49\linewidth}
		\centering
		\includegraphics[scale=\temp]{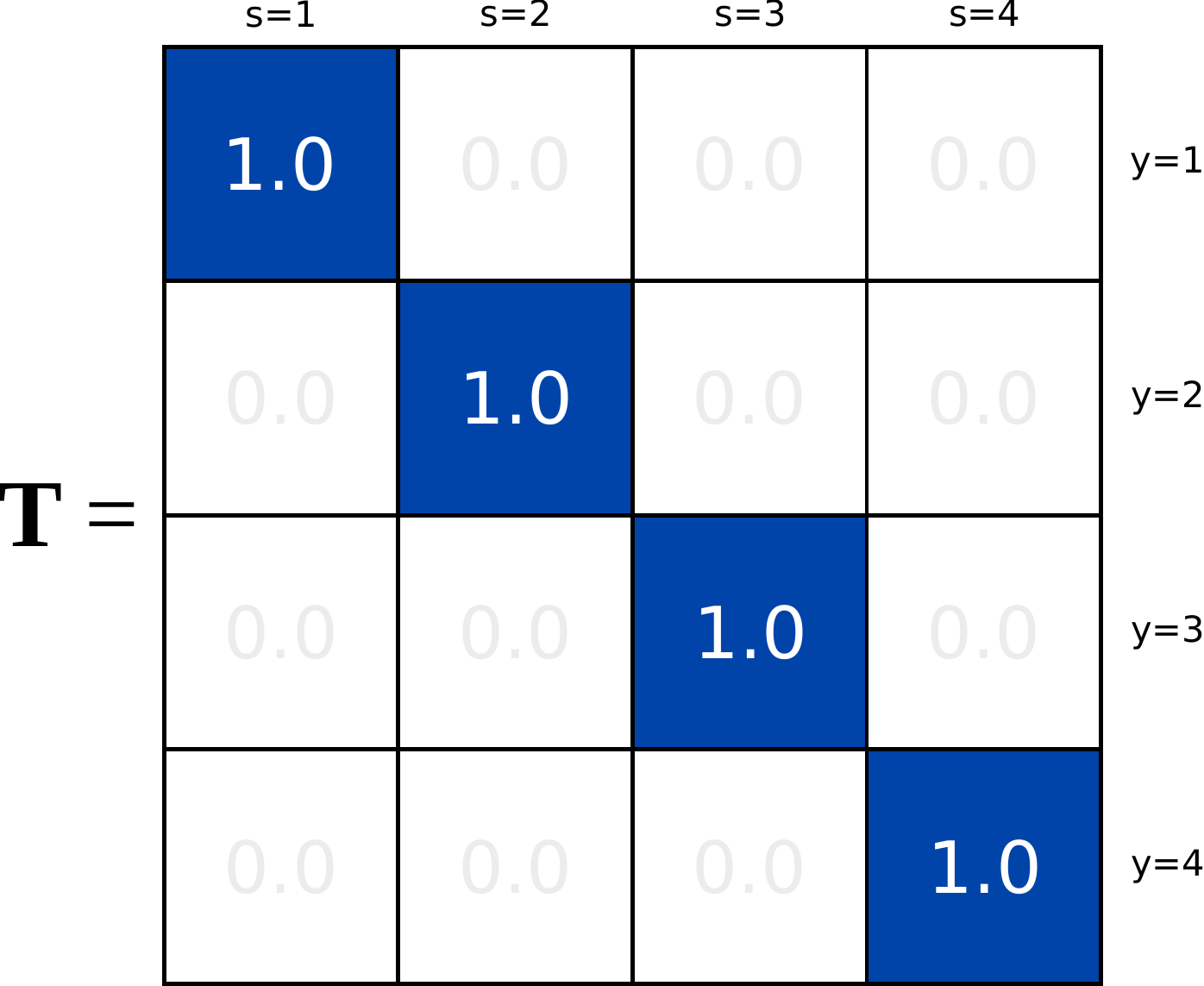}
		\caption{}
	\end{subfigure}
	\begin{subfigure}{0.49\linewidth}
		\centering
		\includegraphics[scale=\temp]{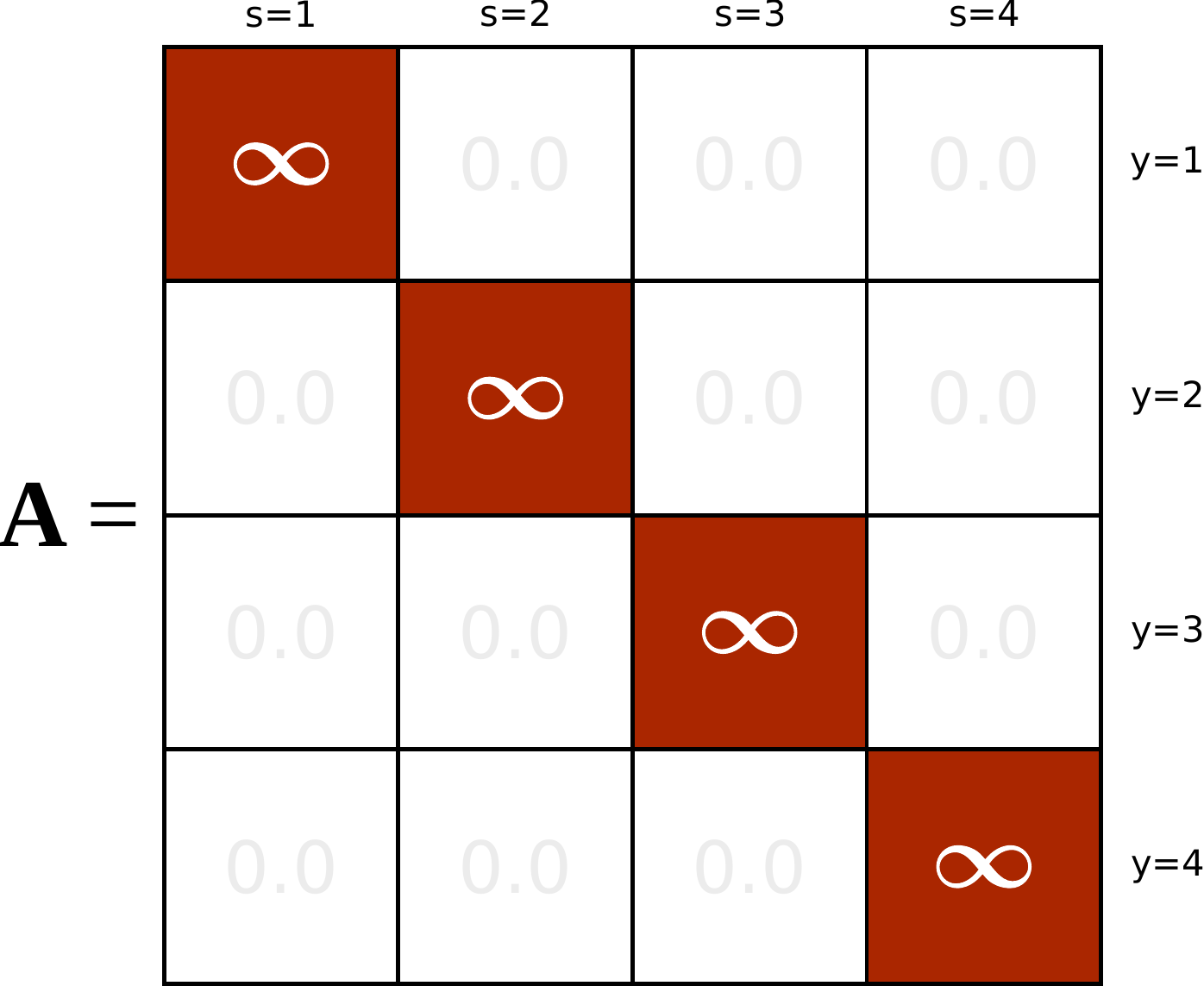}
		\caption{}
	\end{subfigure}
	\caption{Figure with $\mT$'s and $\mA$'s for supervised learning - binary and multiclass.} \label{append_fig:special_case_super}
\end{figure}

Using the inference system for this problem is quite excessive, but consider the ELBO likelihood term for the supervised setting
\begin{align}
	\E_q[ &\ln p(\mS \given \mY, \mT) ] \\
		&= 
		\sum_{is} \emS_{is} 
		\int_{q\triangle} 
		\ln \Big( \sum_{y} \emY_{iy} \emT_{ys} \Big) 
		\d \mT \d \mY \nonumber \\
		&= \sum_{is} \emS_{is} 
		\int_{q\triangle} 
		\ln \emY_{is}
		\d \mY 
		= \sum_{is} \emS_{is} \E[ \ln \emY_{is} ], \nonumber
\end{align}
where we use that the distribution over $\mT$ is a delta-function and that there is a one-to-one correspondence between $y$'s ans $s$'s. Since the class distributions for each sample are independent we can consider a single one in isolation
\begin{align}
	\E_q[ \ln p(\mS_{i:} \given \mY_{i:}, \mT) ] 
		&= \sum_{s} \emS_{is} \E[ \ln \emY_{is} ].
\end{align}
We can select the maximum likelihood solution by solving the following constrained optimization problem
\begin{align}
	\amax_{\vx} \sum_{s} &\emS_{is} \ln(\evx_s)
		= \amax_{\vx} \; \emS_{i:} \bln(\vx), \\
	\vone^{\top}& \vx = 1, 
	\qquad \vx = \emY_{i:}. \nonumber
\end{align}

We make the following Lagrangian with derivatives
\begin{align}
	\lagrange 
		&= \emS_{i:} \bln(\vx) + \lambda (\vone^{\top} \vx - 1), \\
	\diff{\lambda} \lagrange &= \vone^{\top} \vx - 1, \qquad
	\nabla \lagrange = \emS_{i:} \frac{1}{\vx} + \lambda. \nonumber
\end{align}

Setting the last line to zero we find
\begin{gather}
	\emS_{i:} \frac{1}{\vx} + \lambda = 0 \qquad \Leftrightarrow \qquad
	\vx = \frac{1}{\lambda} \emS_{i:}.
\end{gather}

Since the sum of each side of the equation sign has to be equal to 1, we find that 
\begin{align}
	\lambda = \vone^{\top} \emS_{i:} = 1,
\end{align}
because the probabilities of selection labels also sum to 1. 
The optimal solution for the class distributions are therefore simply the label probabilities
\begin{align}
	\mY_{i:}^* = \emS_{i:}.
\end{align}

The inference problem therefore degrades for supervised learning, so that the maximum likelihood solution are simple the label probabilities and the maximum posterior solution will balance this with some term from the class prior.

\subsection{Semi-Supervised Learning}  \label{sec:case_semi_supervised}

Here we expand to the semi-supervised learning problem. In this problem each class will either transition to its own dedicated label or to the unlabelled-label. The transition matrix will now have a diagonal-matrix part, which maps classes to their dedicated labels, as well as a column specifying the transition of being unlabelled from each class (see figure \ref{append_fig:special_case_semisuper}). The distribution over transition matrices will be made to indicate our belief in how much of the respective classes that has been labelled. \\

\renewcommand{\temp}{0.22}
\begin{figure}[h!]
	\centering
	\begin{subfigure}{0.45\linewidth}
		\centering
		\includegraphics[scale=\temp]{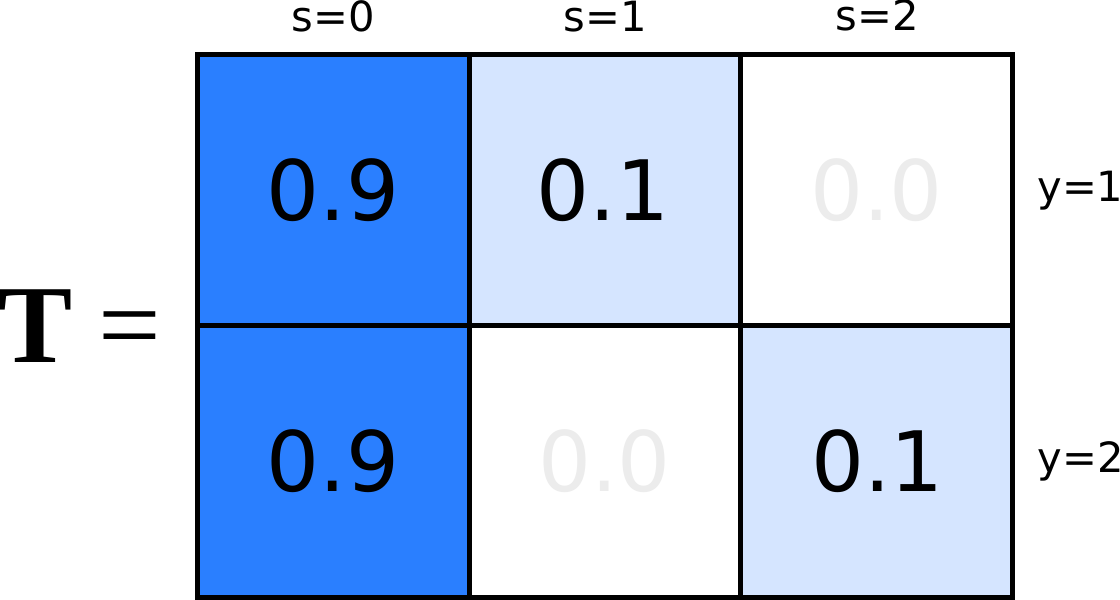}
		\caption{}
	\end{subfigure}
	\begin{subfigure}{0.45\linewidth}
		\centering
		\includegraphics[scale=\temp]{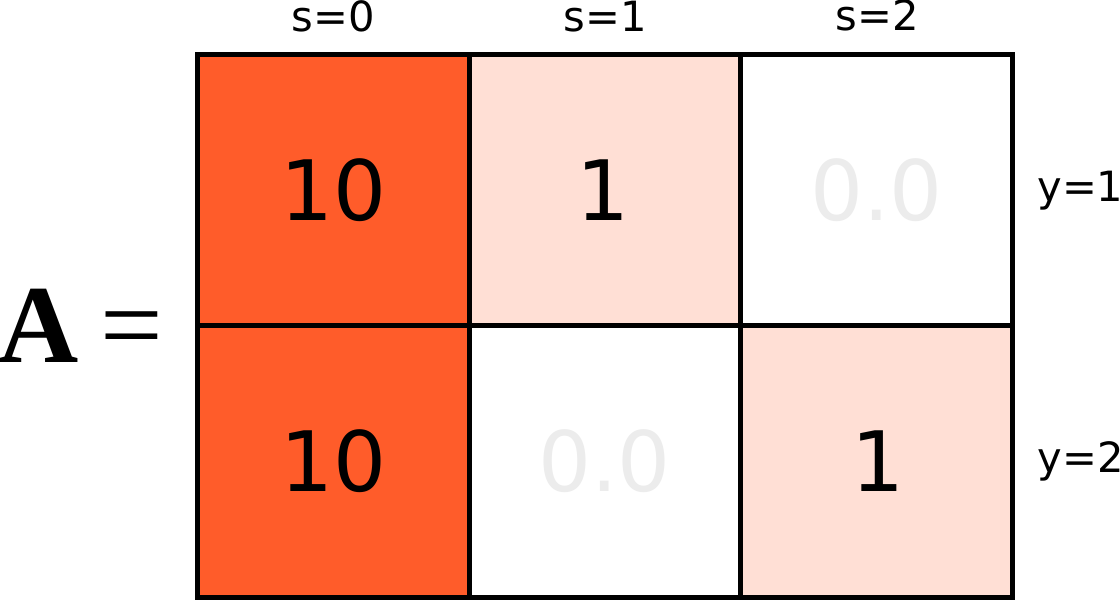}
		\caption{}
	\end{subfigure} \\[5mm]
	\begin{subfigure}{0.49\linewidth}
		\centering
		\includegraphics[scale=\temp]{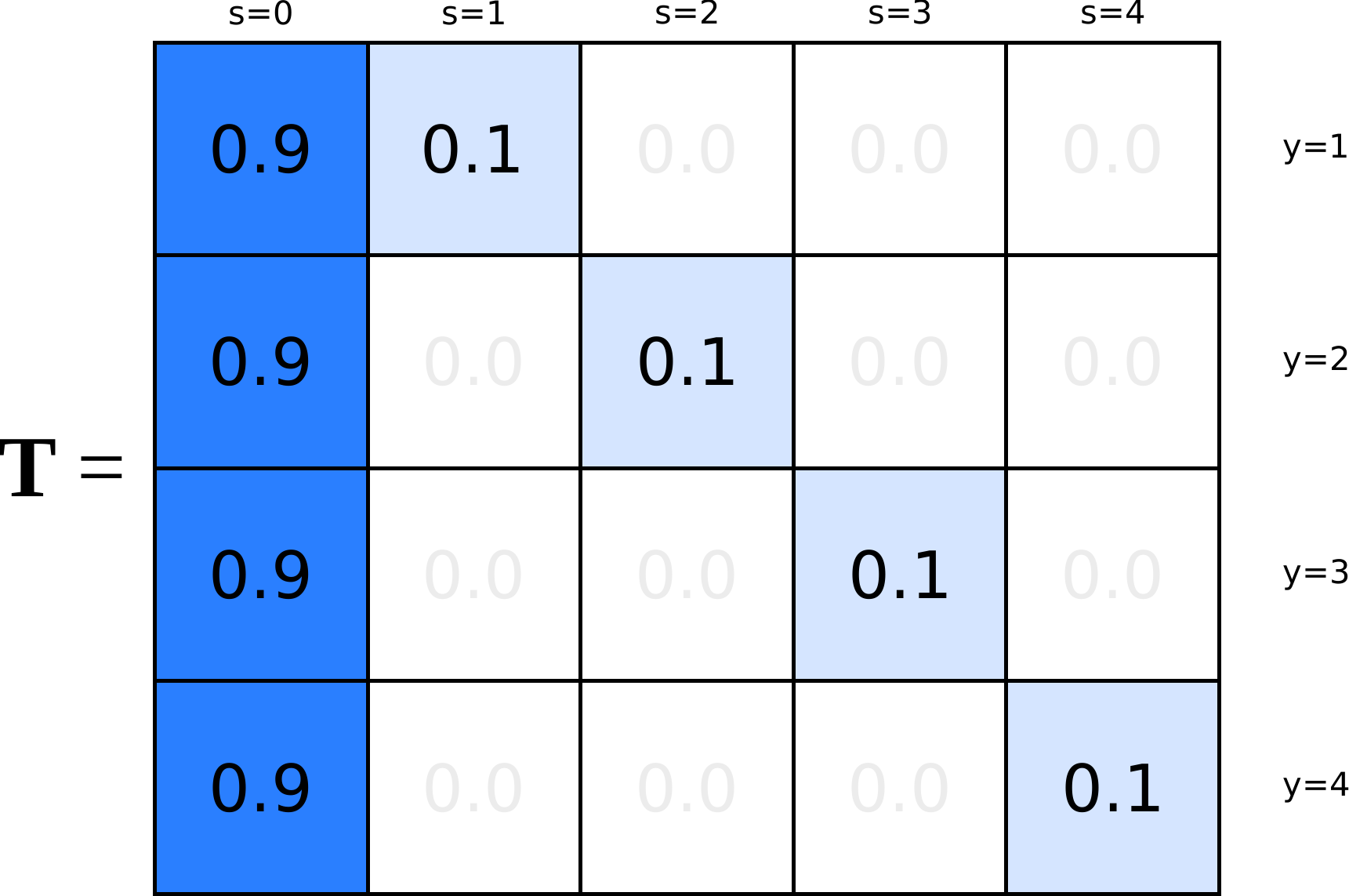}
		\caption{}
	\end{subfigure}
	\begin{subfigure}{0.49\linewidth}
		\centering
		\includegraphics[scale=\temp]{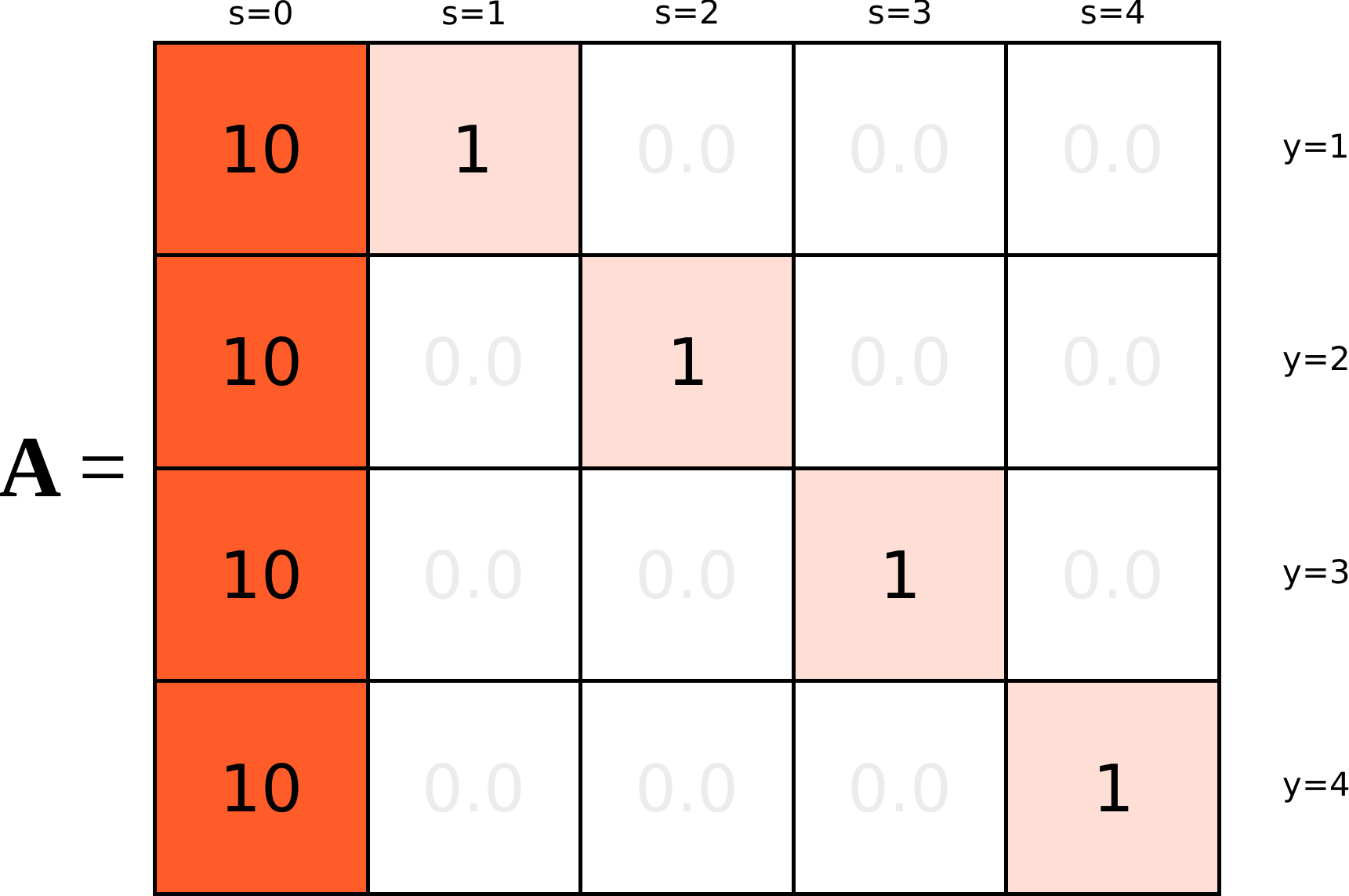}
		\caption{}
	\end{subfigure}
	\caption{Figure with $\mT$'s and $\mA$'s for semi-supervised learning - binary and multiclass.} 
	\label{append_fig:special_case_semisuper}
\end{figure}

Let's consider the likelihood term of a single sample from the ELBO
\begin{align}
	\E_q[ &\ln p(\mS_{i:} \given \mY_{i:}, \mT) ] \\ 
		&= 
		\sum_{s} \emS_{is} 
		\int_{q\triangle} 
		\ln \Big( \sum_{y} \emY_{iy} \emT_{ys} \Big) 
		\d \mT \d \mY. \nonumber
\end{align}

The integral becomes different depending on what selection we are considering. We let $s=0$ be the unlabelled-label, while $s \in \N$ are the labels that correspond to the classes which use the same indexes.
The likelihood term becomes
\begin{align}
	\E_q[ &\ln p(\mS_{i:} \given \mY_{i:}, \mT) ] 
		= 
		\sum_{s=1}^{m_s} \emS_{is} 
		\E\big[ \ln ( \emY_{is} \emT_{ss} ) \big] \\
		&+
		\emS_{i0} \E\Big[ \ln \Big( \sum_{y} \emY_{iy} \emT_{y0} \Big) \Big]. \nonumber
\end{align}

\subsubsection{Equal Probability for Being Unlabelled}
Sometimes it may be valid to assume that the probability of a sample being unlabelled is equal for all classes. That is
\begin{align}
	\forall (i, j) &\in \N, \emT_{i0} = \emT_{j0} = p(u) \\
	\forall (i, j) &\in \N, \emT_{ii} = \emT_{jj} = 1 - \emT_{i0} = 1 - \emT_{j0} = p(\ell), \nonumber
\end{align}
where we have introduced a shorthand notation for these probabilities. \\

In this case we can evaluate the integral above
\begin{align}
	\E_q[ &\ln p(\mS_{i:} \given \mY_{i:}, \mT) ] \\
		&= 
		\sum_{s=1}^{m_s} \emS_{is} 
		\E\big[ \ln ( \emY_{is} p(\ell) ) \big] \nonumber \\
		&\qquad+
		\emS_{i0} \E\Big[ \ln \Big( p(u) \sum_{y} \emY_{iy} \Big) \Big] \nonumber \\
		&= 
		\sum_{s=1}^{m_s} \emS_{is} 
		\Big( 
			\E\big[ \ln p(\ell) \big] 
			+ \E\big[ \ln \emY_{is} \big]
		\Big) \nonumber \\
		&\qquad+
		\emS_{i0} \Big( 
			\E\big[ \ln p(u) \big] 
			+ \E\big[ \ln \big(\vone^{\top} \emY_{i:}\big) \big] 
		\Big) \nonumber \\
		&= 
		\E\big[ \ln p(\ell) \big] (1 - \emS_{i0})
		+ \sum_{s=1}^{m_s} \emS_{is} \E\big[ \ln \emY_{is} \big] \nonumber \\
		&\qquad+ \emS_{i0} \E\big[ \ln p(u) \big]. \nonumber
\end{align}

In a similar manner to that of the supervised case we can find the maximum-likelihood solution by
\begin{align}
	\amax_{\vx}& \sum_{s=1}^{m_s} \emS_{is} \ln(\evx_s)
			= \amax_{\vx} \; \vs^{\top} \bln(\vx), \\
		\vone^{\top} &\vx < 1, 
		\qquad \vx = \begin{bmatrix}
			\emY_{i1} \\ \vdots \\ \emY_{im_s}
		\end{bmatrix}, 
		\quad \vs = \begin{bmatrix}
			\emS_{i1} \\ \vdots \\ \emS_{im_s}
		\end{bmatrix}. \nonumber
\end{align}
The optimal point for this problem is clearly where $\vone^{\top} \vx = 1$. The Lagrangian becomes
\begin{align}
	\lagrange 
		&= \vs^{\top} \bln(\vx) + \lambda (\vone^{\top} \vx - 1), \\
	\diff{\lambda} \lagrange &= \vone^{\top} \vx - 1, \qquad
	\diff{\evx_s} \lagrange = \vs \frac{1}{\evx_i} + \lambda. \nonumber
\end{align}

Setting the last line to zero we find
\begin{gather}
	\vs \frac{1}{\vx} + \lambda = 0 \qquad \Leftrightarrow \qquad
	\vx = \frac{1}{\lambda} \vs.
\end{gather}

As we want $\vx$ to sum to one we find that
\begin{align}
	\lambda = \vone^{\top} \vs = 1 - \emS_{i0}.
\end{align}

The optimal choice for the class probabilities therefore becomes the renormalized label-probabilities
\begin{align}
	\mY_{i:}^* = \frac{1}{1 - \emS_{i0}} \begin{bmatrix}
		\emS_{i1} \\ \vdots \\ \emS_{im_s}
	\end{bmatrix}.
\end{align}

This is as expected, since the best guess for the classes must necessarily be the labels. As the unlabelled probability $\emS_{i0}$ increases we expect the distribution of the class probabilities to widen and become less certain, while keeping its expectation around the point $\mY_{i:}^*$.

\section{Mathematics}

%

\subsection{Gamma Function}  \label{append:gamma_function}

The gamma function is defined by the factorial
\begin{align}
	\Gamma(x) = (x - 1)!, \label{eq:def_gamma_function}
\end{align}
usually interpreted with integers. \\

An alternative definition which does not require integers is
\begin{align}
	\Gamma(x) = \int_0^{\infty} z^{x - 1} \e^{-z} \d z,
\end{align}
which is defined for all complex numbers.

\subsection{Polygamma Functions}  \label{append:polygamma_function}

We define the digamma function as the following auxiliary function for the derivative of the gamma function
\begin{align}
	\digamma(x) = 
		\diff{x} \ln \Gamma(x) = \frac{\Gamma'(x)}{\Gamma(x)}. \label{eq:def_digamma_function}
\end{align}

The digamma function alternatively has an integral definition of
\begin{align}
	\digamma(x) = 
		\int_0^{\infty}\Big( \frac{\e^{-t}}{t} - \frac{\e^{-xt}}{1 - \e^{-t}} \Big) \d t.
\end{align}

The general definition for poly-gamma functions is
\begin{align}
	\digamma^{(m)}(x) = \diff[^m]{x^m} \digamma(x) = \diff[^{m+1}]{x^{m+1}} \ln \Gamma(x),
		\label{eq:def_polygamma_function}
\end{align}

so that the $0$'th poly-gamma function is the digamma function
\begin{align}
	\digamma^{(0)}(x) = \digamma(x). 
\end{align}

\subsection{F-scores}  \label{append:f_scores}

F-scores (or F-measures) are measures of a test's performance. It is generally defined as
\begin{align}
	F_{\beta} 
		&= (1 + \beta^2) 
		\frac{ 
			\text{Pr} \cdot \text{R} 
		}{
			\beta^2 \cdot \text{Pr} + \text{R}
		} \\
		&= \frac{
			(1 + \beta^2) \cdot \text{TP}
		}{
			(1 + \beta^2) \cdot \text{TP} + 
			\beta^2 \cdot \text{FN} + \text{FP}
		}. \nonumber
\end{align}
where
\begin{align}
	Pr &= \text{precision} \\
	R &= \text{Recall} \\
	TP &= \text{True positives} \\
	FP &= \text{False positives} \\
	FN &= \text{False negatives}
\end{align}

The most well-known F-score is the F1-score
\begin{align}
	F_1 = \frac{
		2 \cdot \text{TP}
	}{
		2 \cdot \text{TP} + 
		\text{FN} + \text{FP}
	}.
\end{align}

We use the \textit{macro}-strategy\footnote{~\parbox[t][][t]{\linewidth}{F1-score in \texttt{sklearn} with averaging strategies at: \\ \url{https://scikit-learn.org/stable/modules/generated/sklearn.metrics.f1_score}}} for averaging the F1-score in multi-class problems, where we average the F1-score on each positive class
\begin{align}
	F_{1\text{avg}} = \frac{1}{|\mathcal{P}|} \sum_{y \in \mathcal{P}} F_1( \mY_{:y} ).
\end{align}

\subsection{Moments}  \label{append:moments}

The expectation $\mu = \E[X]$ of a random variable $X$ is given by
\begin{align}
	\mu = \int_{-\infty}^{\infty} x \cdot p(x) \d x. 
\end{align}

The $n$th \textit{central} moment is given by
\begin{align}
	\mu_n = \E\Big[ \big( X - \E[X] \big)^n \Big]. 
\end{align}

The first few central moments are 
\begin{align}
	\mu_0 = \E\Big[ \big( X - \E[X] \big)^0 \Big] &= 1 \\
	\mu_1 = \E\Big[ \big( X - \E[X] \big) \Big] &= 0 \\
	\mu_2 = \E\Big[ \big( X - \E[X] \big)^2 \Big] &= \var[x] \\
	\mu_3 = \E\Big[ \big( X - \E[X] \big)^3 \Big] &= \text{skewness}(x) \\
	\mu_4 = \E\Big[ \big( X - \E[X] \big)^4 \Big] &= \text{kurtosis}(x).
\end{align}

A recursive formula for central moments is
\begin{align}
	\mu_k &= \E\big[ (X - \mu)^k \big] 
		= \sum_{j=0}^k {k \choose j} (-1)^{k - j} \E[X^j] \mu^{k - j}.
\end{align}

\section{Dirichlet Distribution}

A Dirichlet distribution $\dirichlet(\valpha)$ with parameters $\valpha$ is 
\begin{align}
	\valpha \in [0, \infty]^{K},
	&\qquad \bm{X} \sim \dirichlet(\valpha), \\
	\vx \in [0, 1]^{K}, 
	&\qquad p(\vx) = \frac{1}{B(\valpha)} \prod_k^K \evx_k^{\eva_k - 1},
\end{align}
where $B(\cdot)$ is the beta function used for normalizing the distribution.  \\

The parameters for the Dirichlet distribution is $\valpha$ and the sum of the parameters is
\begin{align}
	\evalpha_0 = \sum_{j=1}^K \evalpha_j.
\end{align}

\subsection{Normalization Constant}

The normalization constant (also known as the beta-function) for the Dirichlet distribution is
\begin{align}
	B(\valpha) &= \frac{ \prod_{j=1}^K \Gamma(\evalpha_j) }{ \Gamma(\evalpha_0) } 
		= \frac{ \prod_{j=1}^K \Gamma(\evalpha_j) }{ \Gamma\Big( \sum_{j=1}^K \evalpha_j \Big) } 
		\label{eq:dirichlet_normalization} \\
	\ln B(\valpha) &= \sum_{j=1}^K \ln\Gamma(\evalpha_j) - \ln \Gamma(\evalpha_0), \nonumber
\end{align}
with its logarithm. \\

We will be using the product of normalization constants for normalizing the product of distributions. We define the product of normalization constants as the product of the beta-function on the rows of a matrix
\begin{align}
	B(\mA) = \prod_i B(\mA_{i:}). \label{eq:def_multivar_beta_function_matrix}
\end{align}

The derivative of the log-normalization constant is
\begin{align}
	\diff{\evalpha_i} \ln B(\valpha) 
		&= \diff{\evalpha_i} \ln\Gamma(\evalpha_i) 
		- \diff{\evalpha_i} \ln \Gamma(\evalpha_0) 
		\label{eq:dirichlet_normalization_derivative} \\
		&= \digamma(\evalpha_i) - \digamma(\evalpha_0), \nonumber
\end{align}
using the digamma function (section \ref{append:polygamma_function}).

\subsection{Entropy}

The entropy of the Dirichlet distribution is
\begin{align}
	H(\valpha) &= \ln B(\valpha) + (\evalpha_0 - K) \digamma(\evalpha_0) 
	\label{eq:dirichlet_entropy} \\
		&\qquad- \sum_{j=1}^K (\evalpha_j - 1) \digamma(\evalpha_j). \nonumber
\end{align}

The partial derivative of the entropy if
\begin{align}
	\diff{\evalpha_i}& H(\valpha) \\
		&= \underbrace{\diff{\evalpha_i}\ln B(\valpha)}_1
			+ \underbrace{\diff{\evalpha_i}(\evalpha_0 - K) \digamma(\evalpha_0)}_2
			\label{eq:dirichlet_entropy_derivative} \\
			&\qquad- \underbrace{\diff{\evalpha_i} (\evalpha_i - 1) \digamma(\evalpha_i)}_3 \nonumber \\
		&= \underbrace{\digamma(\evalpha_i) - \digamma(\evalpha_0)}_1 \nonumber \\
			&\qquad+ \underbrace{\digamma(\evalpha_0) + (\evalpha_0 - K) \diff{\evalpha_i} \digamma(\evalpha_0)}_2 \nonumber \\
		    &\qquad- \Big( \underbrace{\digamma(\evalpha_i) + (\evalpha_i - 1) \diff{\evalpha_i} \digamma(\evalpha_i)}_3 \Big) \nonumber \\
		&= (\evalpha_0 - K) \trigamma(\evalpha_0) 
			- (\evalpha_i - 1) \trigamma(\evalpha_i), \nonumber
\end{align}
where the trigamma function $\trigamma(\cdot)$ is the derivative of the digamma function $\digamma(\cdot)$ (section \ref{append:polygamma_function}).

\subsection{Expectation}

The expectation of a Dirichlet variables is
\begin{align}
	\E[ X_j ] = \frac{\evalpha_j}{\evalpha_0}. 
	\label{eq:dirichlet_element_expectation}
\end{align}

The weighted sum of expected elements across a Dirichlet distribution is
\begin{align}
	\sum_{j=0}^K \evw_j \E[ X_j ] 
		= \sum_{j=0}^K \evw_j \frac{\evalpha_j}{\evalpha_0}
		= \frac{1}{\evalpha_0} \sum_{j=0}^K \evw_j \evalpha_j.
		\label{eq:dirichlet_w_sum_expectation}
\end{align}

The partial derivative of the weighted sum of expected elements is
\begin{align}
	\diff{ \evalpha_i } \sum_{j=0}^K \evw_j \E[ X_j ]
		&= \frac{1}{\evalpha_0} \diff{ \evalpha_i } \sum_{j=0}^K \evw_j \evalpha_j 
		\label{eq:dirichlet_w_sum_expectation_derivative} \\
		&= \frac{\evw_i}{\evalpha_0} \diff{ \evalpha_i } \evalpha_i
		= \frac{\evw_i}{\evalpha_0}.
		\nonumber
\end{align}

\subsection{Expected Log}

The expectation of the log of Dirichlet variables is
\begin{align}
	\E[ \ln X_j ] = \digamma(\evalpha_j) - \digamma(\evalpha_0). 
	\label{eq:dirichlet_log_element_expectation}
\end{align}

The weighted sum of expected logs across a Dirichlet distribution is
\begin{equation}
	\sum_{j=0}^K \evw_j \E[ \ln X_j ] 
		= \sum_{j=0}^K \evw_j \digamma(\evalpha_j) 
		- \digamma(\evalpha_0) \sum_j \evw_j.
		\label{eq:dirichlet_w_sum_log_expectation}
\end{equation}

The partial derivative of the weighted sum of expected logs
\begin{align}
	\diff{ \evalpha_i }& \sum_{j=0}^K \evw_j \E[ \ln X_j ] 
		\label{eq:dirichlet_w_sum_log_expectation_derivative} \\
		&= \diff{ \evalpha_i } \digamma(\evalpha_i) \evw_i
			- \diff{ \evalpha_i } \digamma(\evalpha_0) \sum_j \evw_j \nonumber \\
		&= \trigamma(\evalpha_i) \evw_i - \trigamma(\evalpha_0) \sum_j \evw_j. \nonumber
\end{align}

\subsection{Generic Moments}

Consider the generic moment of the Dirichlet distribution
\begin{align}
	\E\Big[ \prod_i^K \vx_i^{\vbeta_i} \Big],
\end{align}
which is a product of any number of the Dirichlet variables to some power. \\

The integral of the expectation is
\begin{align}
	\int_{\dirichlet, \valpha} \prod_i^K &\vx_i^{\vbeta_i} \d \vx \\
		&= \int_{\Delta} \frac{1}{B(\valpha)} \prod_k^K \evx_k^{\eva_k - 1} \prod_i^K \vx_i^{\vbeta_i} \d \vx \nonumber \\
		&= \frac{1}{B(\valpha)} \int_{\Delta} \prod_k^K \evx_k^{\eva_k + \vbeta_k - 1} \d \vx \nonumber \\
		&= \frac{B(\valpha + \vbeta) }{B(\valpha)} \nonumber \\
		&= \frac{ \prod_{j=1}^K \Gamma(\evalpha_j + \evbeta_j) }{ \Gamma(\evalpha_0 + \evbeta_0) } 
			\Bigg( \frac{ \prod_{j=1}^K \Gamma(\evalpha_j) }{ \Gamma(\evalpha_0) }  \Bigg)^{-1} \nonumber
			 \\
		&= 
			\frac{ \prod_{j=1}^K \Gamma(\evalpha_j + \evbeta_j) }{ \Gamma(\evalpha_0 + \evbeta_0) } 
			\frac{ \Gamma(\evalpha_0) }{ \prod_{j=1}^K \Gamma(\evalpha_j) } \nonumber \\
		&= \frac{ \Gamma(\evalpha_0) }{ \Gamma(\evalpha_0 + \evbeta_0) } 
			\frac{ \prod_{j=1}^K \Gamma(\evalpha_j + \evbeta_j) }{ \prod_{j=1}^K \Gamma(\evalpha_j) } \nonumber \\
		&= \frac{ \Gamma\Big( \sum_i^K \evalpha_i \Big) }{ \Gamma\Big( \sum_i^K (\evalpha_i + \evbeta_i) \Big) } \prod_i^K \frac{\Gamma(\evalpha_i + \evbeta_i)}{\Gamma(\evalpha_i)}. \nonumber
\end{align}

Thus the generic moments of the Dirichlet is
\begin{align}
	\E\Big[ \prod_i^K \vx_i^{\vbeta_i} \Big] 
		&= \frac{ \Gamma\Big( \sum_i^K \evalpha_i \Big) }{ \Gamma\Big( \sum_i^K (\evalpha_i + \evbeta_i) \Big) } \prod_i^K \frac{\Gamma(\evalpha_i + \evbeta_i)}{\Gamma(\evalpha_i)}, \nonumber \\
		&\qquad \vbeta \in \big(\R_+ \cup \{0\}\big)^{K}.
		\label{eq:generic_moments_of_dirichlet}
\end{align}

\section{Other Distributions}

\subsection{Categorical Distribution}

A categorical distribution $\categorical(\vp)$ (or single-trial multinomial distribution) with parameters $\vp$ is defined by
\begin{align}
	&\vp \in [0, 1]^K, 
	\qquad \sum_k^K \evp_k = 1, 
	\qquad \bm{X} \sim \categorical(\vp), \nonumber \\
	&\vx \in \Big\{ \vv \in \{0, 1\}^K : \vone^{\top}\vv = 1 \Big\}, 
	\qquad \prod_k^K \evp_k^{\evx_k}.
\end{align}

\subsection{Dirichlet Conjugate Prior Distribution}

The conjugate prior $\Dirp(\vr, \eta)$ for the Dirichlet distribution is described by \cite{andreoli_conjugate_2018}, with parameters $\vr$ and $\eta$ defined by
\begin{align}
	\vr \in \R^{K}, \qquad
	\eta \in [-1, \infty], \\
	\bm{X} \sim \Dirp(\vr, \eta), \qquad
	\vx \in [0, \infty]^K, \nonumber \\
	p(\vx) = \frac{1}{Z(\eta, \vr) B(\vx)^{\eta}} \e^{- \sum_k^K \evr_k \evx_k},\nonumber 
\end{align}
where $B(\cdot)$ is the beta function and $Z(\cdot)$ is a normalizing constant dedicated to this distribution. $Z(\cdot)$ is not easily computed, so for optimization we will often simply work with the unnormalized distribution
\begin{align}
	p(\vx) &\propto \bar{p}(\vx) = \frac{1}{B(\vx)^{\eta}} \e^{- \sum_k^K \evr_k \evx_k} \\
	\ln \bar{p}(\vx) &= -\sum_k^K \evr_k \evx_k - \eta \ln(B(\vx)).
\end{align}
Andreoli \cite{andreoli_conjugate_2018} suggest numerical methods for computing $Z(\cdot)$.

\end{document}